\documentclass[conference]{IEEEtran}
\IEEEoverridecommandlockouts
\usepackage[a-1b]{pdfx}
\usepackage{balance}
\usepackage{hyperref} 
\usepackage{booktabs} 
\usepackage{xcolor}    
\usepackage{colortbl} 
\usepackage{subcaption}
\usepackage{cite}
\usepackage{amsmath,amssymb,amsfonts}
\usepackage{algorithmic}
\usepackage{graphicx}
\usepackage{textcomp}
\usepackage{tabularx}  

\def\BibTeX{{\rm B\kern-.05em{\sc i\kern-.025em b}\kern-.08em
    T\kern-.1667em\lower.7ex\hbox{E}\kern-.125emX}}
\begin{document}

\title{Pre‑Tactical Flight‑Delay and Turnaround Forecasting with Synthetic Aviation Data
\thanks{Preprint. Under review.}
}

\author{
    \IEEEauthorblockN{Abdulmajid Murad$^{1}$, Massimiliano Ruocco$^{1,2}$}
    \IEEEauthorblockA{
        \parbox{0.5\textwidth}{
            \centering
            $^1$Department of Software Engineering, Safety and Security \\ 
            SINTEF Digital \\
            Trondheim, Norway \\
            \{abdulmajid.murad, massimiliano.ruocco\}@sintef.no
        }
        \hspace{0.05\textwidth}
        \parbox{0.4\textwidth}{
            \centering
            $^2$Department of Computer Science \\
            Norwegian University of Science and Technology \\
            Trondheim, Norway \\
            massimiliano.ruocco@ntnu.no
        }
    }
}

\maketitle

\begin{abstract}

    Access to comprehensive flight operations data remains severely restricted in aviation due to commercial sensitivity and competitive considerations, hindering the development of predictive models for operational planning. This paper investigates whether synthetic data can effectively replace real operational data for training machine learning models in pre-tactical aviation scenarios—predictions made hours to days before operations using only scheduled flight information. We evaluate four state-of-the-art synthetic data generators on three prediction tasks: aircraft turnaround time, departure delays, and arrival delays.  
    Using a Train on Synthetic, Test on Real (TSTR) methodology on over 1.7 million European flight records, we first validate synthetic data quality through fidelity assessments, then assess both predictive performance and the preservation of operational relationships.
    Our results show that advanced neural network architectures, specifically transformer-based generators, can retain 94-97\% of real-data predictive performance while maintaining feature importance patterns informative for operational decision-making.
    Our analysis reveals that even with real data, prediction accuracy is inherently limited (R² $\leq$ 0.44) when only scheduled information is available—establishing realistic baselines for pre-tactical forecasting. 
    These findings suggest that high-quality synthetic data can enable broader access to aviation analytics capabilities while preserving commercial confidentiality, though stakeholders must maintain realistic expectations about pre-tactical prediction accuracy given the stochastic nature of flight operations.

\end{abstract}


\section{Introduction}

The aviation industry generates substantial operational data that could be valuable for optimizing flight operations, improving resource allocation, and enhancing passenger experience. However, access to this data is often restricted due to commercial sensitivity, privacy regulations, and competitive considerations. Airlines typically guard their operational data as commercially sensitive and proprietary, concerned that disclosure could affect their competitive strategies.
Additionally, data protection laws such as GDPR and security concerns create tensions between data openness and privacy protection \cite{lei2025fedmeta}. This limited access affects external researchers, smaller airlines, and emerging markets that lack the historical datasets needed for developing predictive analytics.

The challenge of data scarcity is particularly relevant in pre-tactical flight operations—the period hours to days before actual flights when airlines, airports, and passengers must make decisions based on anticipated conditions. Unlike tactical predictions made on the day of operations with real-time updates, pre-tactical models must rely primarily on published schedules and historical patterns \cite{dalmau2024probabilistic}. While tactical models can incorporate live weather updates, air traffic flow restrictions, and current aircraft positions, pre-tactical predictions operate with more limited information where only scheduled characteristics and statistical patterns are available \cite{mas2022pre}.

Despite these information constraints, pre-tactical predictions can provide value across the aviation ecosystem. Airlines use these forecasts to adjust operations—modifying aircraft assignments, crew schedules, or flight times before disruptions propagate through their networks \cite{mas2022pre}. Airports utilize predictions to plan gate requirements and staffing needs based on expected delays and turnaround times \cite{de2023probabilistic}. For passengers, advance notice of potential delays can inform decisions about connections and travel arrangements \cite{carvalho2021relevance}. However, organizations with limited access to historical data—including smaller airlines, regional airports, and academic researchers—face challenges in developing these predictive capabilities.

The uncertainty in pre-tactical prediction stems from various factors that influence actual operations but remain unknown at schedule publication time. Weather conditions, air traffic congestion, technical issues, crew availability, airport capacity constraints, and cascading delays from earlier flights all contribute to deviations from scheduled operations \cite{carvalho2021relevance}. For aircraft turnaround time—the period between arrival and subsequent departure—predictions must account for ground processes including passenger handling, cleaning, catering, fueling, and maintenance, each varying by aircraft type, airport infrastructure, and time of day \cite{de2023probabilistic}. Similarly, departure and arrival delay predictions must infer potential disruptions based on historical patterns and scheduled characteristics, without access to current operational status.

This operational complexity presents two related challenges. First, even with comprehensive data access, pre-tactical predictions face accuracy limitations due to the stochastic nature of aviation operations \cite{dalmau2024probabilistic}. Weather patterns, technical failures, and human factors introduce uncertainties that are difficult to capture days in advance. Second, the restricted availability of operational data prevents many organizations from developing baseline predictive models. Without historical records showing how scheduled operations translated into actual performance under various conditions, stakeholders cannot build fundamental predictive capabilities.

Recent work in synthetic data generation presents one approach to addressing data availability while respecting commercial confidentiality. By creating artificial datasets that preserve statistical properties and predictive relationships of real operations without exposing sensitive information, synthetic data could enable broader access to analytics capabilities. The Train on Synthetic, Test on Real (TSTR) paradigm \cite{esteban2017real} suggests that models trained on synthetic data can generalize to real-world scenarios, potentially allowing organizations without historical data access to develop operational prediction capabilities. Recent work has shown that synthetic flight data can approximate real data's statistical profiles, with models trained on synthetic data achieving prediction accuracies comparable to those trained on actual data \cite{aly2025synthetic}.

However, the effectiveness of synthetic data for pre-tactical aviation prediction requires further investigation. The challenges of pre-tactical scenarios—where predictive relationships are embedded in complex interactions between limited scheduled features—require evaluation beyond traditional distributional similarity metrics. An important consideration is whether synthetic data can preserve not only the statistical properties of flight operations but also the latent patterns that enable prediction of future delays and extended turnarounds from scheduled information.

This study evaluates the utility of synthetic flight data across three pre-tactical prediction tasks: aircraft turnaround time, departure delays, and arrival delays. We assess four synthetic data generation approaches—Gaussian Copula (statistical modeling), Conditional Tabular GAN (adversarial learning) \cite{xu2019modeling}, TabSyn (diffusion-based synthesis) \cite{zhang2024mixed}, and REaLTabFormer (transformer architecture) \cite{solatorio2023realtabformer}—examining their effectiveness in preserving relationships necessary for operational forecasting when only scheduled information is available.

Our evaluation examines synthetic data quality through fidelity metrics, predictive performance comparisons, and feature relationship preservation. This ensures that models trained on synthetic data achieve acceptable accuracy while identifying the same operational drivers as real-data models—critical for practical deployment where understanding prediction rationale can be as valuable as the predictions themselves

Using over 1.7 million European flight records, our analysis contributes:
\begin{itemize}
\item An evaluation framework for assessing synthetic data utility in pre-tactical aviation contexts, accounting for limited feature availability and operational uncertainty.
\item A comprehensive fidelity evaluation framework spanning five dimensions—statistical similarity, correlation preservation, joint distribution fidelity, likelihood-based assessment, and detection difficulty—to validate synthetic aviation data quality before operational deployment.
\item Quantitative analysis of performance differences when substituting synthetic for real training data across multiple operational prediction tasks, finding that transformer-based methods can retain 94-97\% of real-data predictive performance.
\item Identification of synthetic generation approaches that preserve predictive relationships when operational information is limited, with REaLTabFormer demonstrating consistent performance in maintaining feature importance patterns.
\item An open-source implementation of the used generative models and evaluation pipeline, making the methods accessible for further research and benchmarking.\footnote{Code available at: \href{https://github.com/SynthAIr/syntabair}{https://github.com/SynthAIr/syntabair}}
\end{itemize}

Through this analysis, we explore whether synthetic data can serve as a practical substitute for proprietary operational records in pre-tactical prediction tasks. Our findings suggest potential pathways for broader access to analytics capabilities across the industry while maintaining commercial confidentiality requirements.

\section{Related Work}

The challenge of predicting flight delays and turnaround times has attracted significant attention in recent years. De Falco et al. \cite{de2023probabilistic} developed probabilistic machine learning models to forecast aircraft turnaround times and Target Off-Block Times (TOBT) at major European airports. This work uniquely focused on turnaround operations, which are critical contributors to reactionary delays, and provided probabilistic outputs rather than point estimates.

In the domain of delay prediction, Mas-Pujol et al. \cite{mas2022pre} presented a machine learning framework to predict individual flight delays. Their model employs a two-stage approach: first classifying whether flights will experience delays and identifying the cause category, then predicting delay duration. The study identified crucial features including flight timing, airport demand, and weather conditions that influence pre-tactical delays, demonstrating the advantages of data-driven approaches over traditional simulation methods.

Dalmau et al. \cite{dalmau2024probabilistic} introduced probabilistic delay prediction using quantile regression. Their work predicts delay distributions rather than single estimates, enabling better uncertainty quantification. Notably, their analysis revealed that expected passenger numbers significantly influenced delay predictions - an insight often unavailable in other studies due to data limitations.

A comprehensive survey by Wandelt et al. \cite{wandelt2025flight}  examined approximately 40 recent flight delay prediction studies. Their analysis revealed that machine learning approaches, particularly random forests and neural networks, have become dominant in capturing delay factors. Common features across studies include scheduled information, weather conditions, and route characteristics, highlighting the consistent importance of weather and operational timing in delay forecasting.

The proprietary nature of aviation datasets has driven researchers toward synthetic data generation as an alternative approach. Most existing work has concentrated on trajectory data generation. Krauth et al. \cite{krauth2023deep}, Ezzahed et al. \cite{ezzahed2022bringing}, and Krauth et al. \cite{krauth2024advanced} demonstrated the effectiveness of Variational Autoencoders (VAEs) for modeling aircraft trajectories in terminal maneuvering areas. Wijnands et al. \cite{wijnands2024generation} explored Generative Adversarial Networks (GANs), specifically the TimeGAN framework, for creating synthetic landing trajectories. Murad and Ruocco \cite{murad2025synthetic} extended this trajectory-focused work by applying vector-quantized VAEs to generate synthetic flight paths between airport pairs. Their results showed that synthetic trajectories could effectively capture the statistical properties of real flight data, though these approaches remain limited to spatial trajectory modeling rather than operational records. 

A notable departure from trajectory-focused work comes from Aly and Sharpanskykh \cite{aly2025synthetic}, who broadened synthetic data generation to encompass flight operational records. Their study generated tabular datasets including flight schedules, aircraft information, and delay data. However, their evaluation was limited to two generative models (TVAE and Gaussian Copula) and used publicly available U.S. domestic flight data.

While substantial progress has been made in both flight delay prediction and synthetic data generation, a critical gap remains in combining these approaches for pre-tactical operational planning. Existing delay prediction models rely on access to comprehensive historical datasets, which many smaller operators and researchers lack. Similarly, synthetic data generation in aviation has primarily focused on trajectories rather than the operational records needed for delay prediction. This work addresses this gap by investigating whether synthetic operational data can enable effective pre-tactical delay prediction without requiring access to sensitive historical records.

\section{Methodology}

\subsection{Overview} 
\begin{figure*}[t]
    \centering
    \includegraphics[width=\textwidth]{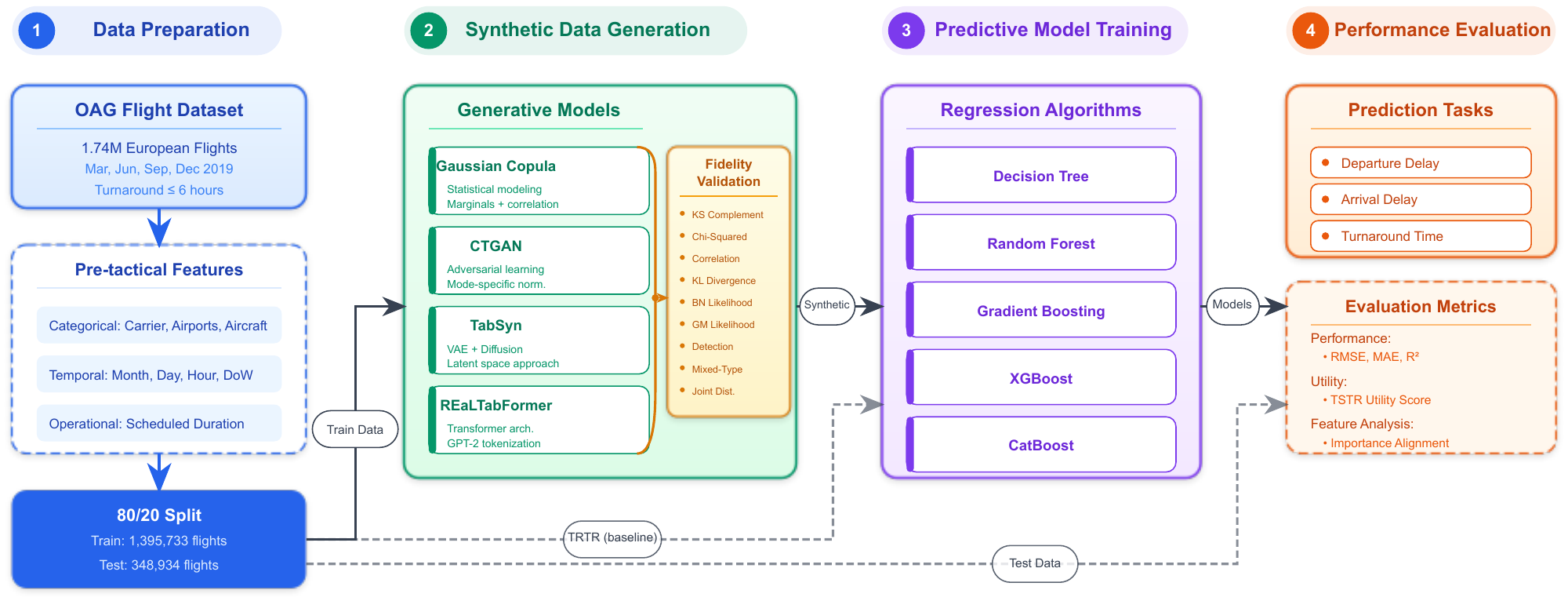}
    \caption{Experimental framework comprising four phases: (1) Data Preparation - processing 1.74M European flight records and extracting pre-tactical features available at schedule publication; (2) Synthetic Data Generation - training four generative models and validating statistical properties; (3) Predictive Model Training - implementing parallel pipelines for baseline (TRTR) and synthetic data evaluation (TSTR) using five regression algorithms; (4) Performance Evaluation - assessing performance on three operational prediction tasks using multiple metrics.}
    \label{fig:framework}
\end{figure*}

Our methodology, illustrated in Figure \ref{fig:framework}, follows a four-phase experimental framework designed to systematically evaluate synthetic data utility for pre-tactical flight prediction.

In Phase 1 (Data Preparation), we process 1.74 million flight records from the OAG dataset, extracting only features available at schedule publication time—categorical variables (carrier, airports, aircraft type), temporal features (month, day, hour, day of week), and operational metrics (scheduled duration). The dataset is split 80/20 for training and testing, with turnaround times constrained to six hours to exclude overnight parking.

Phase 2 (Synthetic Data Generation) involves training four generative models on the real training data: Gaussian Copula (statistical modeling), CTGAN (adversarial learning), TabSyn (VAE with diffusion), and REaLTabFormer (transformer architecture). Before proceeding to model training, we conduct comprehensive fidelity assessment using statistical similarity metrics, correlation preservation tests, joint distribution analysis, and detection difficulty measures to ensure synthetic data quality.

Phase 3 (Predictive Model Training) establishes two parallel pipelines. The baseline pipeline (Train on Real, Test on Real - TRTR) trains five regression models—Decision Tree, Random Forest, Gradient Boosting, XGBoost, and CatBoost—on real training data. The evaluation pipeline (Train on Synthetic, Test on Real - TSTR) trains the same models on each synthetic dataset, enabling direct comparison of predictive performance.

Phase 4 (Peformance Evaluation) assesses all trained models on the held-out real test set across three pre-tactical prediction tasks: departure delay, arrival delay, and turnaround time. We employ performance metrics (RMSE, MAE, R²), utility scores quantifying synthetic-to-real performance ratios, and feature importance alignment to ensure models identify the same operational drivers regardless of training data source.

\subsection{Phase 1: Data Preparation}

\subsubsection{Flight Dataset}

We utilized OAG's Flight Info Direct Status Summary database\footnote{'Status Direct: Data Fields Explained'. Accessed: June 17, 2025. [Online]. Available: \url{https://knowledge.oag.com/docs/status-direct-data-fields-explained}}, extracting European flight operations from March, June, September, and December 2019. After filtering for completed flights with full operational timelines and valid turnaround observations, we obtained 1,744,667 flight records.

The turnaround matching process ensured all records have valid turnaround observations, enabling consistent evaluation across all three prediction tasks. The dataset exhibits mean departure delay of 11.1 minutes ($\sigma$=24.0), mean arrival delay of 7.5 minutes ($\sigma$=25.8), and mean turnaround time of 70.8 minutes ($\sigma$=51.2).

We applied an 80/20 train-test split (1,395,733/348,934 flights) with fixed random seed for reproducibility.

\subsubsection{Feature Selection}
For pre-tactical prediction, we selected only features available at schedule publication time:

\begin{itemize}
    \item Categorical: IATA carrier code, departure/arrival airports, aircraft type
    \item Temporal: Month, day, hour, minute (extracted from scheduled departure), day of week
    \item Operational: Scheduled flight duration (actual duration excluded in pre-tactical mode)
\end{itemize}

\subsubsection{Target Variables}
We computed three operational metrics:
\begin{itemize}
    \item Departure delay: $\Delta_{\text{dep}} = t_{\text{AOBT}} - t_{\text{SOBT}}$ (Actual vs. Scheduled Off-Block Time)
    \item Arrival delay: $\Delta_{\text{arr}} = t_{\text{AIBT}} - t_{\text{SIBT}}$ (Actual vs. Scheduled In-Block Time)
    \item Turnaround time: Ground time between consecutive flights of the same aircraft at the same airport, constrained to $\leq$ 6 hours to exclude overnight parking
\end{itemize}

\subsection{Phase 2: Synthetic Data Generation}

\subsubsection{Generative Models}

Four state-of-the-art synthetic data generators were evaluated, each representing different technical approaches to capturing the complex relationships in flight data. Table~\ref{tab:generator_comparison} summarizes their key characteristics.

\begin{table*}[htbp]
\centering
\caption{Comparison of Synthetic Data Generation Models}
\label{tab:generator_comparison}
\small
\begin{tabularx}{\textwidth}{l l X X X}
\toprule
\textbf{Model} & \textbf{Approach} & \textbf{Key Innovation} & \textbf{Strengths} & \textbf{Limitations} \\
\midrule
\textbf{Gaussian Copula} & Statistical & Separates marginals from dependence structure & Fast training, preserves marginal distributions, minimal parameters & Limited to linear correlations, struggles with rare events \\
\midrule
\textbf{CTGAN} & Adversarial (GAN) & Mode-specific normalization for mixed data & Handles multimodal distributions, conditional generation & Training instability, computational cost \\
\midrule
\textbf{TabSyn} & VAE + Diffusion & Two-stage latent space generation & Efficient sampling (20-50 steps), captures complex dependencies & Two-stage complexity, no conditional generation \\
\midrule
\textbf{REaLTabFormer} & Transformer & Row-wise autoregressive generation & Superior relationship preservation, domain constraints & High computational requirements, needs large datasets \\
\bottomrule
\end{tabularx}
\end{table*}

\textbf{Gaussian Copula (GC)} represents the simplest approach, using classical statistical theory to model dependencies between flight variables. For aviation data, it fits appropriate marginal distributions to each feature (e.g., log-normal for delays, categorical for airports) while capturing their correlation structure through a multivariate Gaussian copula. While computationally efficient and requiring minimal training data, its reliance on linear correlations limits its ability to capture the complex non-linear relationships common in flight operations, such as cascading delays or airport-specific congestion patterns.

\textbf{Conditional Tabular GAN (CTGAN)} \cite{xu2019modeling} addresses aviation data's mixed types through specialized preprocessing. Its mode-specific normalization handles the multimodal nature of continuous features like delays (which cluster around zero but have long tails), while training-by-sampling ensures adequate representation of rare categories. The conditional generation capability proves valuable for creating targeted scenarios, though the adversarial training can be unstable and computationally intensive for large flight datasets.

\textbf{TabSyn} \cite{zhang2024mixed} employs two-stage approach suited to aviation's heterogeneous data. First, a variational autoencoder learns continuous representations that capture relationships between categorical features (carriers, airports) and continuous metrics (delays, durations). Then, a diffusion model in this latent space enables efficient generation requiring only 20-50 denoising steps—orders of magnitude faster than traditional diffusion. This efficiency makes it practical for generating large synthetic flight datasets, though the two-stage architecture adds implementation complexity.

\textbf{REaLTabFormer} \cite{solatorio2023realtabformer} leverages transformer architectures to model each flight record as a sequence of tokens. This approach captures the sequential dependencies in flight data—how airline choice influences aircraft type, which affects turnaround time, which impacts subsequent delays. The model's column-aware tokenization and domain-specific constraints ensure generated values remain operationally valid. While achieving the highest fidelity among evaluated methods, it requires substantial computational resources and training data.

For detailed architectural specifications, training procedures, and implementation details of each model, see Appendix~\ref{sec:appendix-generative_models}.

\subsubsection{Fidelity Evaluation}

To ensure synthetic data maintains the statistical properties necessary for reliable downstream analysis, we employ a comprehensive fidelity evaluation framework spanning five complementary dimensions. Each dimension captures different aspects of how well synthetic data reproduces the characteristics of real flight operations.

\textbf{Statistical Similarity}:
We assess univariate distributional fidelity using metrics tailored to data type. For continuous variables (e.g., delays, durations), we employ the Kolmogorov-Smirnov (KS) complement:
\begin{equation}
\text{KS}_{\text{complement}} = 1 - D_{KS}
\end{equation}
where $D_{KS}$ represents the maximum distance between empirical cumulative distribution functions. Values approaching 1.0 indicate near-perfect alignment of continuous distributions. For categorical variables (e.g., carriers, airports), we apply Pearson's Chi-Squared test, converting p-values to a normalized 0-1 scale where higher values signify better frequency matching between real and synthetic categorical distributions.

\textbf{Correlation Preservation}:
Maintaining inter-feature relationships is crucial for operational validity. We evaluate four correlation metrics:
\begin{itemize}
    \item Pearson correlation: Captures linear relationships between continuous variables
    \item Spearman and Kendall correlations: Assess rank-based and ordinal dependencies, respectively
    \item Correlation Matrix Distance: Measures preservation of global correlation structure using Frobenius norm
    \item Mixed-Type correlation: Extends analysis across heterogeneous variable pairs using appropriate measures (Pearson's r for numerical-numerical, Cramér's V for categorical-categorical, correlation ratio for mixed pairs)
\end{itemize}

\textbf{Joint Distribution Fidelity}:
Beyond pairwise relationships, we evaluate multivariate distributional alignment through Kullback-Leibler (KL) divergence. For continuous variables, we transform KL divergence to a fidelity score:
\begin{equation}
\text{Fidelity}_{\text{continuous}} = \exp(-\text{KL}(p||q))
\end{equation}
where $p$ and $q$ represent real and synthetic distributions. We apply analogous measures for discrete variables, assessing how well generators capture complex categorical combinations (e.g., airline-airport pairs).

\textbf{Likelihood-Based Assessment}:
We fit probabilistic models to real data and measure the likelihood of synthetic samples under these learned distributions:
\begin{itemize}
    \item Bayesian Network likelihood: Evaluates categorical dependencies using directed graphical models learned via the Chow-Liu algorithm
    \item Gaussian Mixture Model likelihood: Assesses continuous feature distributions as weighted combinations of multivariate Gaussians
\end{itemize}
Higher log-likelihood values indicate synthetic data that better conforms to the statistical patterns in real operations.

\textbf{Detection Difficulty}:
As a holistic measure of synthetic data quality, we train regularized logistic regression classifiers to distinguish real from synthetic records. The detection score is normalized as:
\begin{equation}
\text{Detection Score} = 1 - (\text{AUC} - 0.5) \times 2
\end{equation}
where AUC represents the area under the ROC curve. Scores approaching 1.0 indicate synthetic data that is statistically indistinguishable from real data, while lower scores suggest detectable artifacts that may impact downstream applications.

This multi-faceted evaluation ensures synthetic data not only mimics surface-level statistics but preserves the complex interdependencies essential for aviation operational analysis.

\subsection{Phase 3: Predictive Model Training}

\subsubsection{Regression Algorithms}
We employed five regression algorithms to ensure robust evaluation across different modeling paradigms:
\begin{itemize}
    \item \textbf{Decision Tree}: Captures non-linear patterns through recursive partitioning
    \item \textbf{Random Forest}: Ensemble of trees for improved generalization
    \item \textbf{Gradient Boosting}: Sequential ensemble focusing on residual errors
    \item \textbf{XGBoost}: Optimized gradient boosting with regularization
    \item \textbf{CatBoost}: Gradient boosting specialized for categorical features
\end{itemize}

\subsubsection{Training Protocols}
\begin{enumerate}
    \item \textbf{Train on Real, Test on Real (TRTR)}: Baseline performance using historical data
    \item \textbf{Train on Synthetic, Test on Real (TSTR)}: Utility evaluation of synthetic data
    \item \textbf{Feature Engineering}: All models used identical pre-tactical features to ensure fair comparison
    \item \textbf{Hyperparameter Optimization}: Default parameters were used to avoid biasing results toward specific generators
\end{enumerate}

\subsection{Phase 4: Performance Evaluation}

\subsubsection{Prediction Tasks}

We evaluated three operational prediction scenarios, all in pre-tactical mode where only scheduled information is available:

\textbf{Turnaround Time Prediction}: 
Forecasting the duration between aircraft arrival (In-Block) and departure (Off-Block) using:
\begin{itemize}
    \item Input features: Scheduled times, airports, carrier, aircraft type, temporal features
    \item Target: Actual turnaround time in minutes
    \item Operational relevance: Critical for gate allocation and ground resource planning
\end{itemize}

\textbf{Departure Delay Prediction}:
Estimating the difference between scheduled and actual departure times:
\begin{itemize}
    \item Input features: Same as turnaround prediction
    \item Target: Departure delay in minutes (positive for late departures)
    \item Operational relevance: Enables proactive delay mitigation strategies
\end{itemize}

\textbf{Arrival Delay Prediction}:
Forecasting arrival delays using only pre-departure information:
\begin{itemize}
    \item Input features: Same as above predictions
    \item Target: Arrival delay in minutes
    \item Challenge: Most difficult task due to cumulative uncertainties from departure delays and en-route factors
\end{itemize}

\subsubsection{Evaluation Metrics}

Our evaluation employed multiple complementary metrics to assess synthetic data quality comprehensively:

\textbf{Predictive Performance Metrics}:
\begin{itemize}
    \item Root Mean Squared Error (RMSE): Emphasizes large prediction errors critical for disruption management:
    \begin{equation}
        \text{RMSE} = \bigl( \tfrac{1}{n}\sum_{i=1}^{n}(y_i-\hat y_i)^2 \bigr)^{1/2}
    \end{equation}
    
    \item Mean Absolute Error (MAE): Provides typical error magnitude for operational planning:
    \begin{equation}
    \text{MAE} = \frac{1}{n}\sum_{i=1}^{n}|y_i - \hat{y}_i|
    \end{equation}
    
    \item Coefficient of Determination (R²): Measures explained variance to assess predictability limits:
    \begin{equation}
    R^2 = 1 - \frac{\sum_{i=1}^{n}(y_i - \hat{y}_i)^2}{\sum_{i=1}^{n}(y_i - \bar{y})^2}
    \end{equation}
\end{itemize}

\textbf{Utility Score}: A normalized metric combining RMSE and R² performance to quantify how well synthetic-trained models substitute for real-trained models:
\begin{subequations}\label{eq:utility}
\begin{IEEEeqnarray}{rCl}
\text{U}_{\text{RMSE}} &=& 
   \min\!\Bigl(
   \tfrac{\text{RMSE}_{\text{real}}}{\text{RMSE}_{\text{syn}}},\,1
   \Bigr)
   \label{eq:utility_rmse}\\[2pt]
\text{U}_{R^{2}} &=& 
   \min\!\Bigl(
   \tfrac{R^{2}_{\text{syn}}}{R^{2}_{\text{real}}},\,1
   \Bigr)
   \label{eq:utility_r2}\\[2pt]
\text{U}_{\text{model}} &=& 
   \tfrac{1}{2}\bigl(
   \text{U}_{\text{RMSE}} + \text{U}_{R^{2}}
   \bigr)
   \label{eq:utility_model}\\[2pt]
\text{U}_{\text{generator}} &=& 
   \tfrac{1}{n}\sum_{i=1}^{n}
   \text{U}_{\text{model},i}
   \label{eq:utility_avg}
\end{IEEEeqnarray}
\end{subequations}

\textbf{Feature Importance Alignment}:
To ensure synthetic data preserves operational relationships, we compute the cosine similarity between feature importance vectors from models trained on real ($\mathbf{v}_{\text{real}}$) versus synthetic ($\mathbf{v}_{\text{synthetic}}$) data:
\begin{equation}
\text{Alignment} = \frac{\mathbf{v}_{\text{real}} \cdot \mathbf{v}_{\text{synthetic}}}{||\mathbf{v}_{\text{real}}|| \cdot ||\mathbf{v}_{\text{synthetic}}||}
\end{equation}

\section{Results and Discussion}

\subsection{Synthetic Data Fidelity}
\label{sec:fidelity_results}

Before evaluating predictive utility, we assessed the statistical fidelity of each synthetic data generator to ensure the generated data maintains essential properties of real flight operations. Table~\ref{tab:fidelity_metrics} presents twelve complementary metrics measuring different aspects of synthetic data quality.

\begin{table}[!htbp]
\centering
\scriptsize
\setlength{\tabcolsep}{1pt}
\renewcommand{\arraystretch}{1.2}
\caption{Synthetic Data Fidelity Metrics}
\label{tab:fidelity_metrics}
\begin{tabular}{lcccc}
\toprule
\rowcolor{blue!15}
\textbf{Metric} & \textbf{GaussianCopula} & \textbf{CTGAN} & \textbf{TabSyn} & \textbf{REaLTabFormer} \\
\midrule
\rowcolor{gray!5}
\textbf{KS Complement} & 0.8779 & 0.9528 & 0.9855 & \textbf{0.9911} \\
\rowcolor{gray!20}
\textbf{Chi-Squared Test} & 0.9979 & 1.0000 & 1.0000 & \textbf{1.0000} \\
\rowcolor{gray!5}
\textbf{Pearson Correlation} & \textbf{0.9955} & 0.9881 & 0.9925 & 0.9950 \\
\rowcolor{gray!20}
\textbf{Spearman Correlation} & 0.9822 & 0.9874 & 0.9962 & \textbf{0.9982} \\
\rowcolor{gray!5}
\textbf{Kendall Correlation} & 0.9865 & 0.9902 & 0.9974 & \textbf{0.9987} \\
\rowcolor{gray!20}
\textbf{Correlation Matrix} & \textbf{0.9896} & 0.9704 & 0.9790 & 0.9887 \\
\rowcolor{gray!5}
\textbf{Mixed-Type Correlation} & 0.9314 & 0.9765 & 0.9854 & \textbf{0.9899} \\
\rowcolor{gray!20}
\textbf{Continuous KL Divergence} & 0.7437 & 0.9709 & 0.9966 & \textbf{0.9990} \\
\rowcolor{gray!5}
\textbf{Discrete KL Divergence} & 0.2272 & 0.6808 & 0.7981 & \textbf{0.9369} \\
\rowcolor{gray!20}
\textbf{BN Log Likelihood} & -18.5829 & -15.0552 & -14.7318 & \textbf{-13.9766} \\
\rowcolor{gray!5}
\textbf{GM Log Likelihood} & -42.3439 & -38.9681 & -37.5899 & \textbf{-37.5268} \\
\rowcolor{gray!20}
\textbf{Logistic Detection} & 0.0253 & 0.6275 & \textbf{0.8880} & 0.8472 \\
\bottomrule
\end{tabular}
\end{table}

The results reveal a clear performance hierarchy across generators. REaLTabFormer demonstrates better fidelity across most metrics, particularly in distributional similarity (KS Complement: 0.991) and joint distribution modeling (Discrete KL Divergence: 0.937). TabSyn follows closely, achieving the highest detection difficulty score (0.888), indicating its synthetic data is most challenging to distinguish from real data. CTGAN shows moderate performance across all dimensions, while Gaussian Copula, despite maintaining accurate marginal distributions and correlations, struggles with complex multivariate relationships as evidenced by low KL divergence scores.

All generators successfully preserve correlation structures ($>$0.97 across Pearson, Spearman, and Kendall metrics), which is crucial for maintaining predictive relationships. The high Chi-Squared scores ($\geq$0.998) indicate accurate reproduction of categorical feature distributions across all methods. These fidelity results confirm that the synthetic data maintains sufficient statistical integrity for the subsequent predictive utility evaluation, with REaLTabFormer and TabSyn emerging as the most faithful generators overall.

\subsection{Departure Delay Prediction}

\begin{figure}[htbp]
    \centering
    \includegraphics[width=\linewidth]{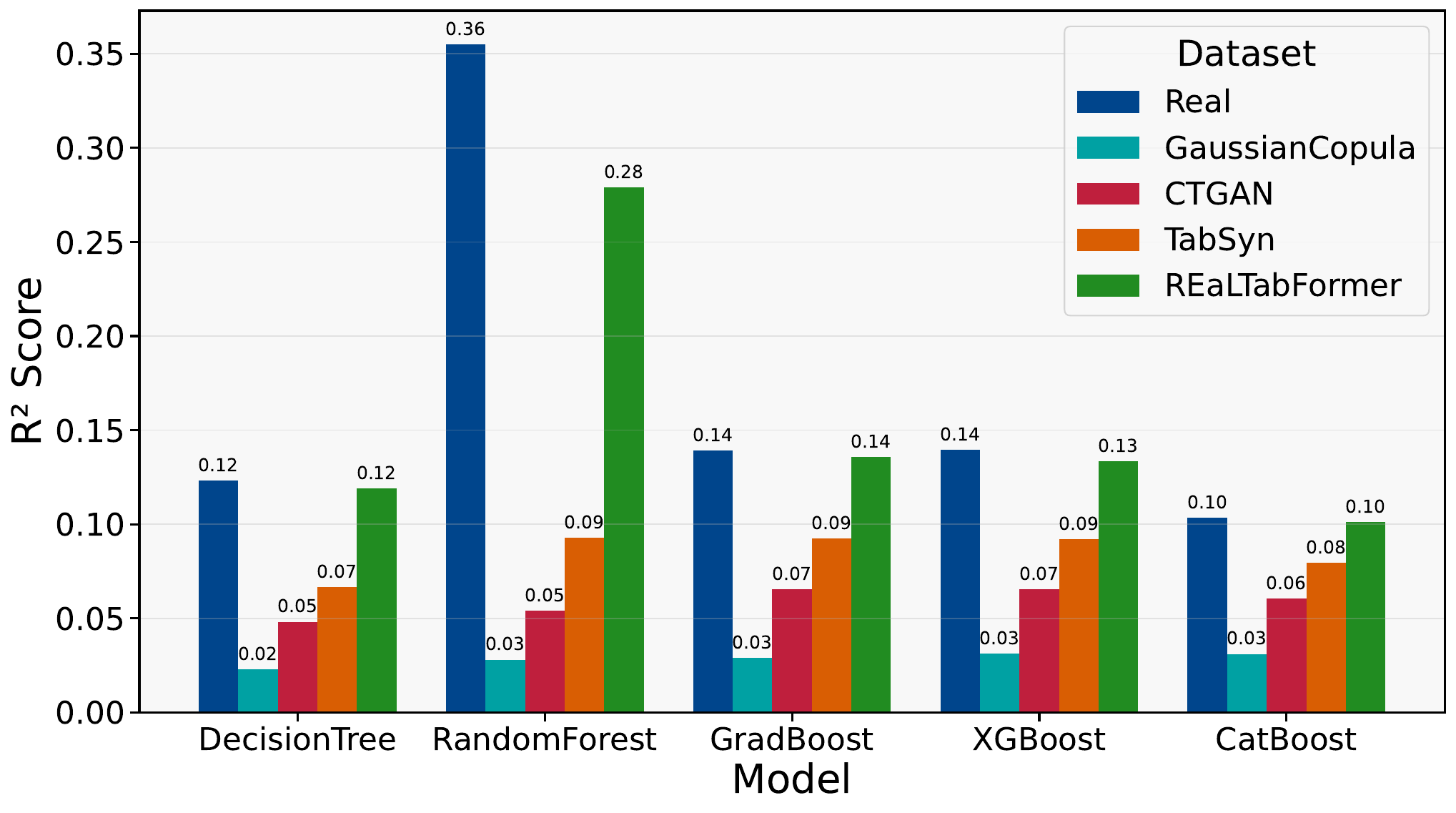}
    \caption{Coefficient of determination (R²) for pre-tactical departure delay prediction across different models and synthetic data generators.}
    \label{fig:departure_r2}
\end{figure}

Pre-tactical departure delay prediction emerged as one of the challenging tasks, with real-data models achieving R² values between 0.10 and 0.30 (Figure~\ref{fig:departure_r2}). This limited predictability reflects the numerous external factors influencing departure times—weather conditions, air traffic flow restrictions, crew availability, and technical issues—that are not captured in scheduled information alone. The inherent stochasticity of these factors creates a fundamental ceiling on pre-tactical prediction accuracy, regardless of data quality.

\begin{figure}[htbp]
    \centering
    \begin{subfigure}[b]{0.49\textwidth}
        \includegraphics[width=\linewidth]{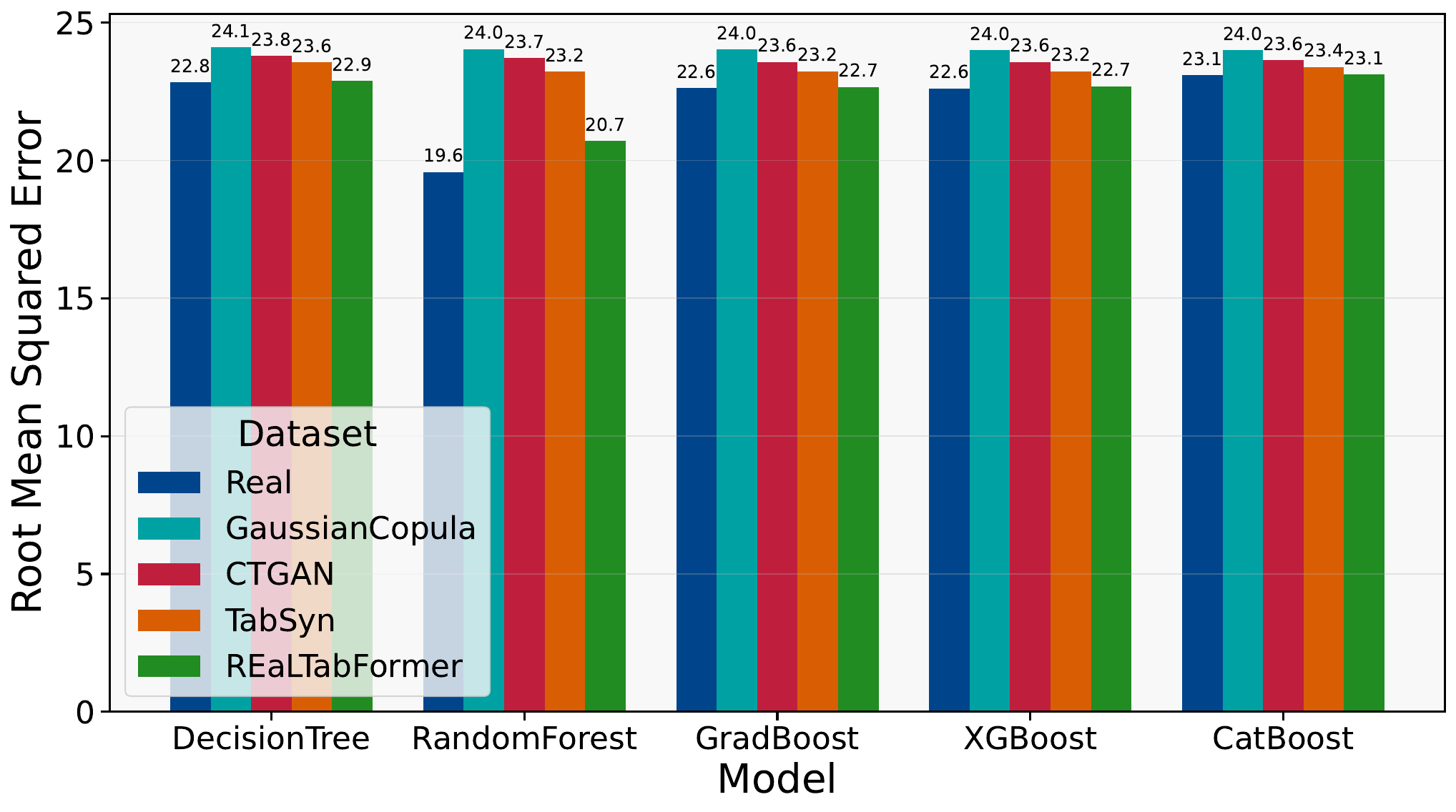}
        \caption{Root Mean Squared Error}
        \label{fig:departure_pre_rmse}
    \end{subfigure}
    \hfill
    \begin{subfigure}[b]{0.49\textwidth}
        \includegraphics[width=\linewidth]{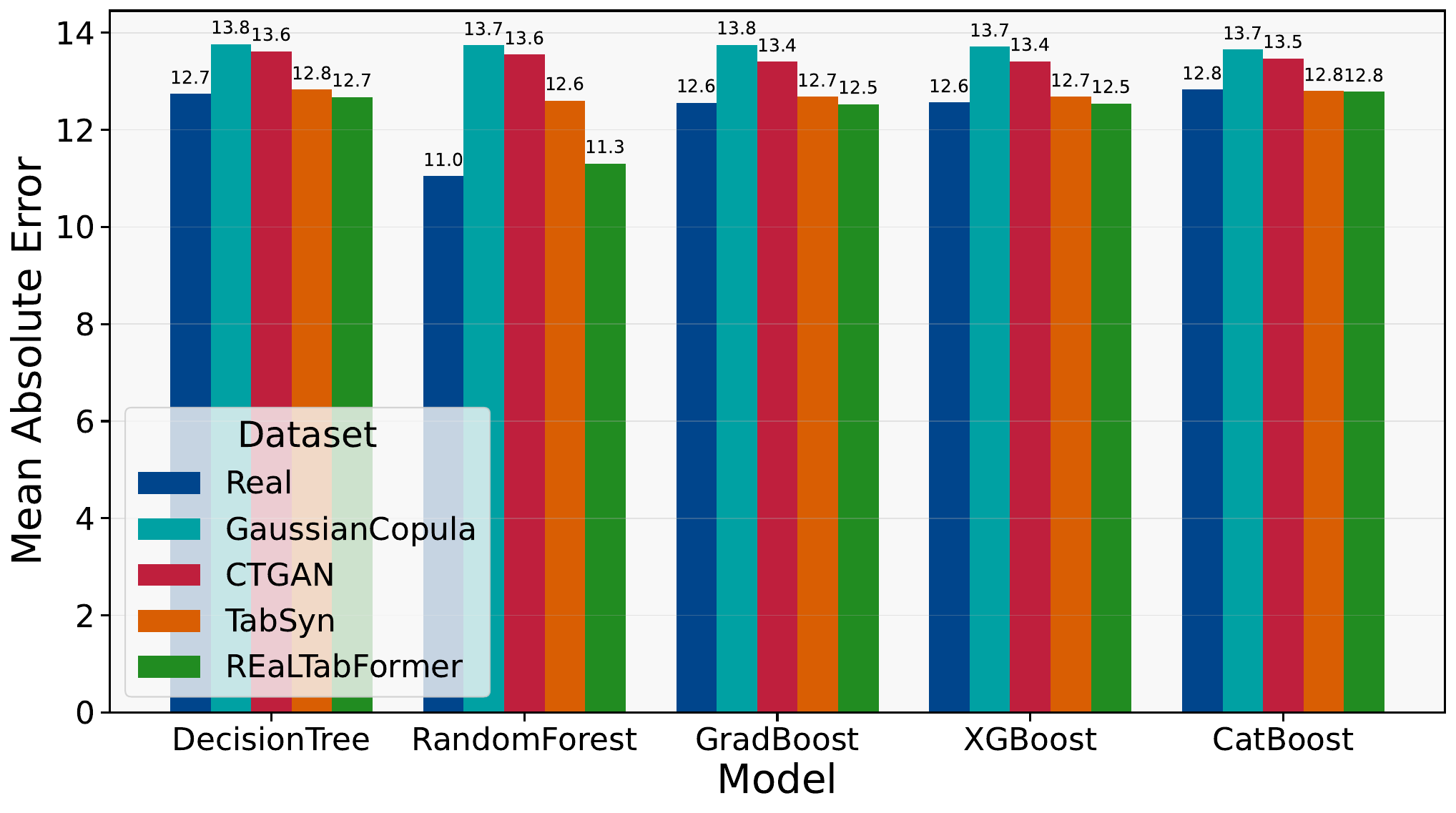}
        \caption{Mean Absolute Error}
        \label{fig:departure_pre_mae}
    \end{subfigure}
    \caption{Prediction error metrics for pre-tactical departure delay across models and synthetic data generators.}
\end{figure}

Despite these constraints, clear performance hierarchies emerged among the synthetic generators. The RMSE and MAE metrics highlight REaLTabFormer's performance advantage in preserving the subtle predictive patterns present in scheduled data (Figures \ref{fig:departure_pre_rmse} and \ref{fig:departure_pre_mae}). REaLTabFormer consistently achieved the lowest error rates, with RMSE values within 4 \% of real-data baselines. The transformer architecture's ability to capture dependencies between categorical features (airlines, airports) and temporal patterns proves crucial for this task.

\begin{figure}[htbp]
    \centering
    \includegraphics[width=0.8\linewidth]{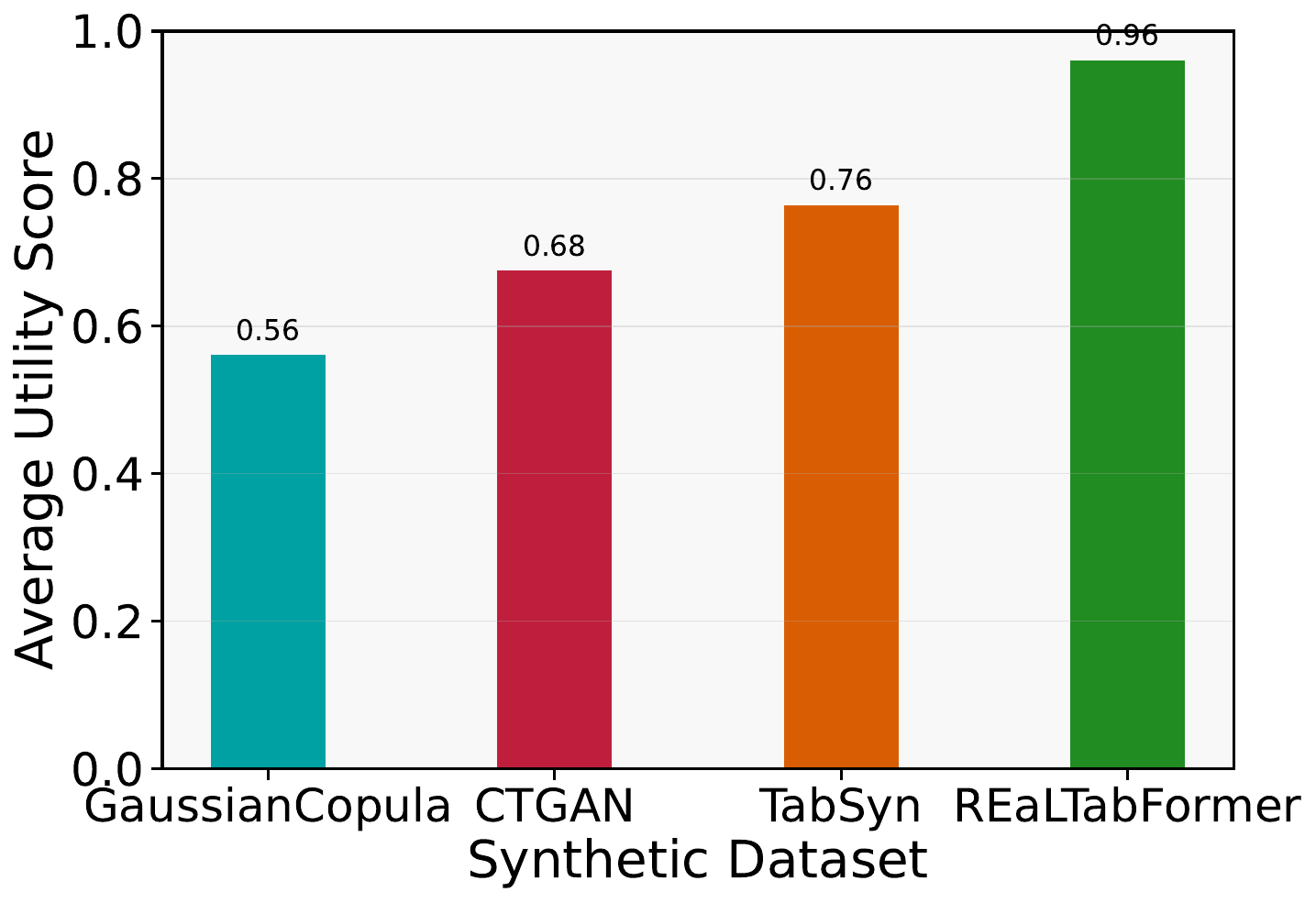}
    \caption{Average utility scores for pre-tactical departure delay prediction across synthetic data generators.}
    \label{fig:departure_utility}
\end{figure}

The utility scores quantify the practical impact of these differences (Figure \ref{fig:departure_utility}). REaLTabFormer achieved a utility score of 0.96, indicating that models trained on its synthetic data retain 96 \% of the predictive capability of models trained on real data. This represents a strong preservation of predictive relationships given the task's inherent difficulty. TabSyn's utility dropped more significantly to 0.76, suggesting that departure-delay patterns require sophisticated modeling of rare events and edge cases that simpler diffusion approaches struggle to capture consistently.

\begin{figure}[htbp]
    \centering
    \includegraphics[width=\linewidth]{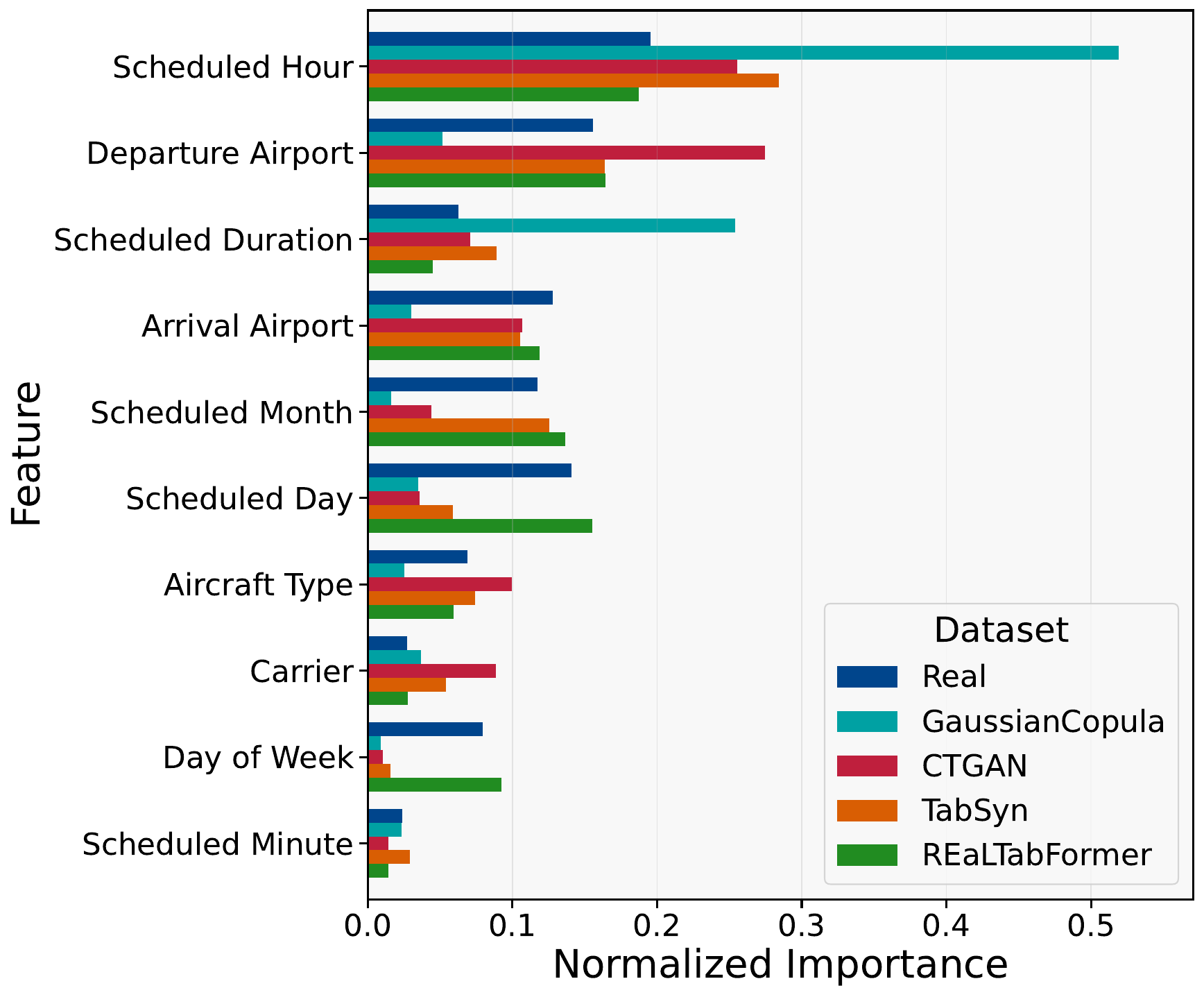}
    \caption{Feature importance comparison for pre-tactical departure delay prediction averaged across all models.}
    \label{fig:departure_features}
\end{figure}

Feature importance analysis reveals that scheduled hour emerged as the dominant predictor (Figure~\ref{fig:departure_features}), capturing well-known operational patterns: morning peak congestion, midday lulls, and evening cascade effects where early delays propagate through the network. Airport features remained important but with reduced relative weight compared to temporal factors, reflecting the systematic nature of delay patterns across different hubs.

\begin{figure}[htbp]
    \centering
    \includegraphics[width=0.8\linewidth]{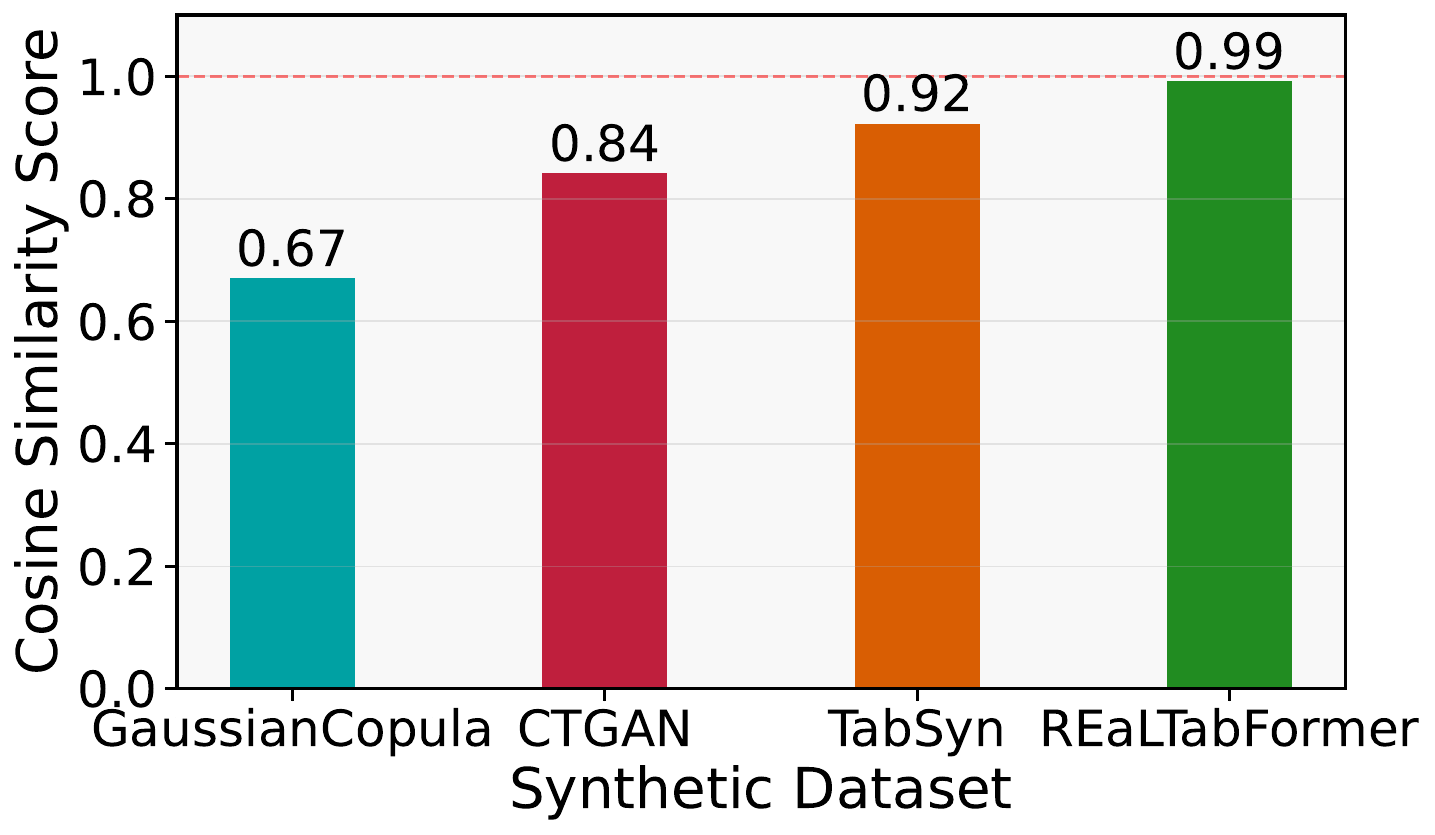}
    \caption{Average feature importance alignment scores for pre-tactical departure delay prediction.}
    \label{fig:departure_alignment}
\end{figure}

The feature alignment analysis reveals interesting patterns (Figure~\ref{fig:departure_alignment}). While REaLTabFormer maintained near-perfect alignment (0.99), indicating that models trained on its synthetic data identify the same operational drivers as real-data models, Gaussian Copula's alignment dropped to 0.67. This misalignment could lead to incorrect operational conclusions if synthetic data users assume the same causal relationships present in real operations.

\subsection{Arrival Delay Prediction}

\begin{figure}[htbp]
    \centering
    \includegraphics[width=\linewidth]{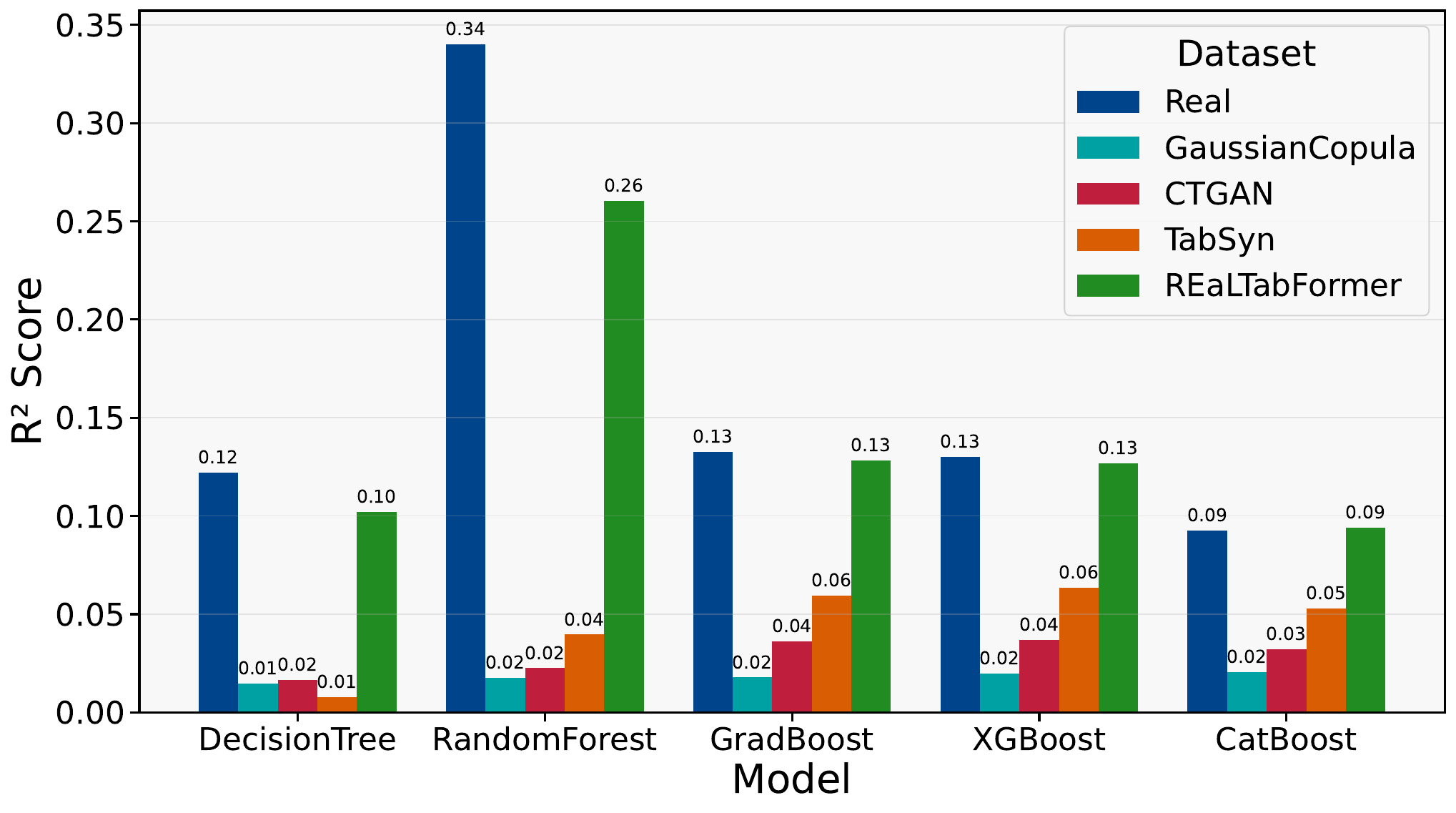}
    \caption{Coefficient of determination (R²) for pre-tactical arrival delay prediction across different models and synthetic data generators.}
    \label{fig:arrival_pre_r2}
\end{figure}

Pre-tactical arrival delay prediction proved to be the most challenging task among all evaluated scenarios, with real-data R² values rarely exceeding 0.30 (Figure~\ref{fig:arrival_pre_r2}). This represents the cumulative uncertainty of multiple operational phases: ground operations affecting departure timing, en-route factors including air traffic congestion and weather, and destination airport conditions. The complexity of predicting arrival delays hours in advance using only scheduled information highlights fundamental limitations in aviation predictability.

\begin{figure}[htbp]
    \centering
    \begin{subfigure}[b]{0.49\textwidth}
        \includegraphics[width=\linewidth]{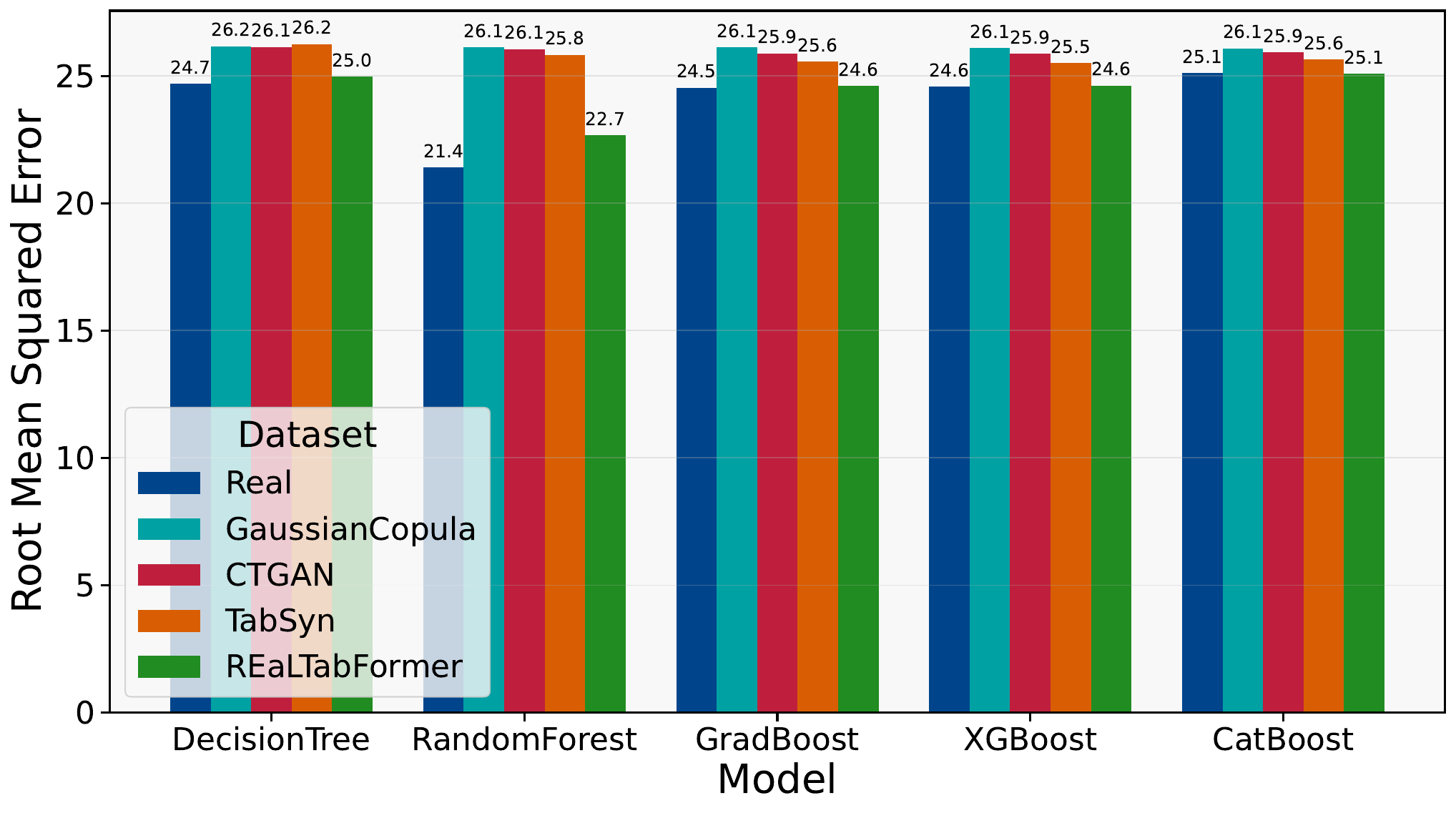}
        \caption{Root Mean Squared Error}
        \label{fig:arrival_pre_rmse}
    \end{subfigure}
    \hfill
    \begin{subfigure}[b]{0.49\textwidth}
        \includegraphics[width=\linewidth]{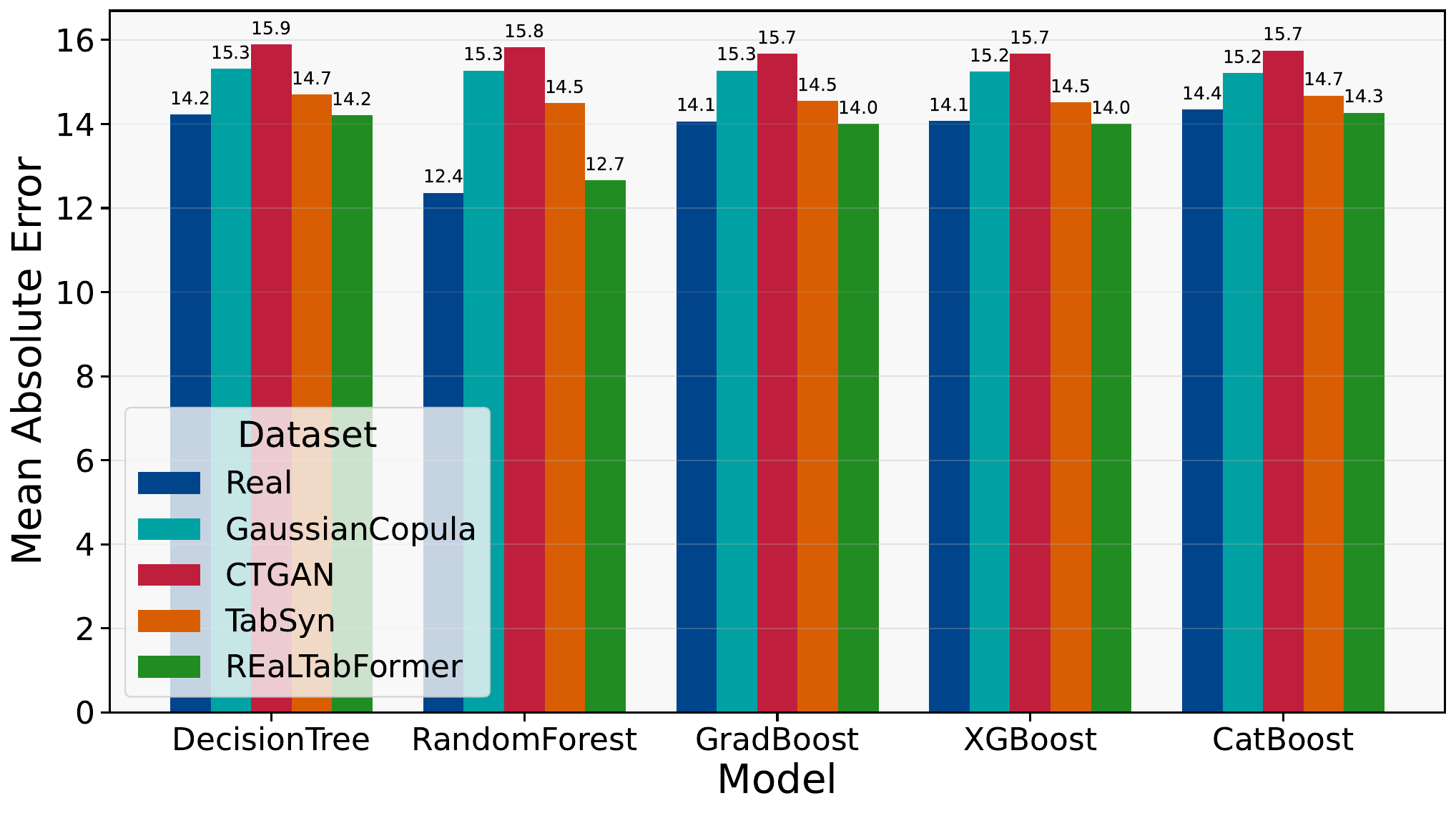}
        \caption{Mean Absolute Error}
        \label{fig:arrival_pre_mae}
    \end{subfigure}
    \caption{Prediction error metrics for pre-tactical arrival delay across models and synthetic data generators.}
\end{figure}

The error metrics demonstrate that even under these challenging conditions, synthetic data generators maintained their relative performance hierarchy (Figures~\ref{fig:arrival_pre_rmse} and \ref{fig:arrival_pre_mae}). REaLTabFormer continued to achieve the lowest error rates. Thus, the transformer architecture's advantage becomes more pronounced in complex scenarios where capturing subtle interactions between multiple categorical and temporal features proves crucial.

\begin{figure}[htbp]
    \centering
    \includegraphics[width=0.8\linewidth]{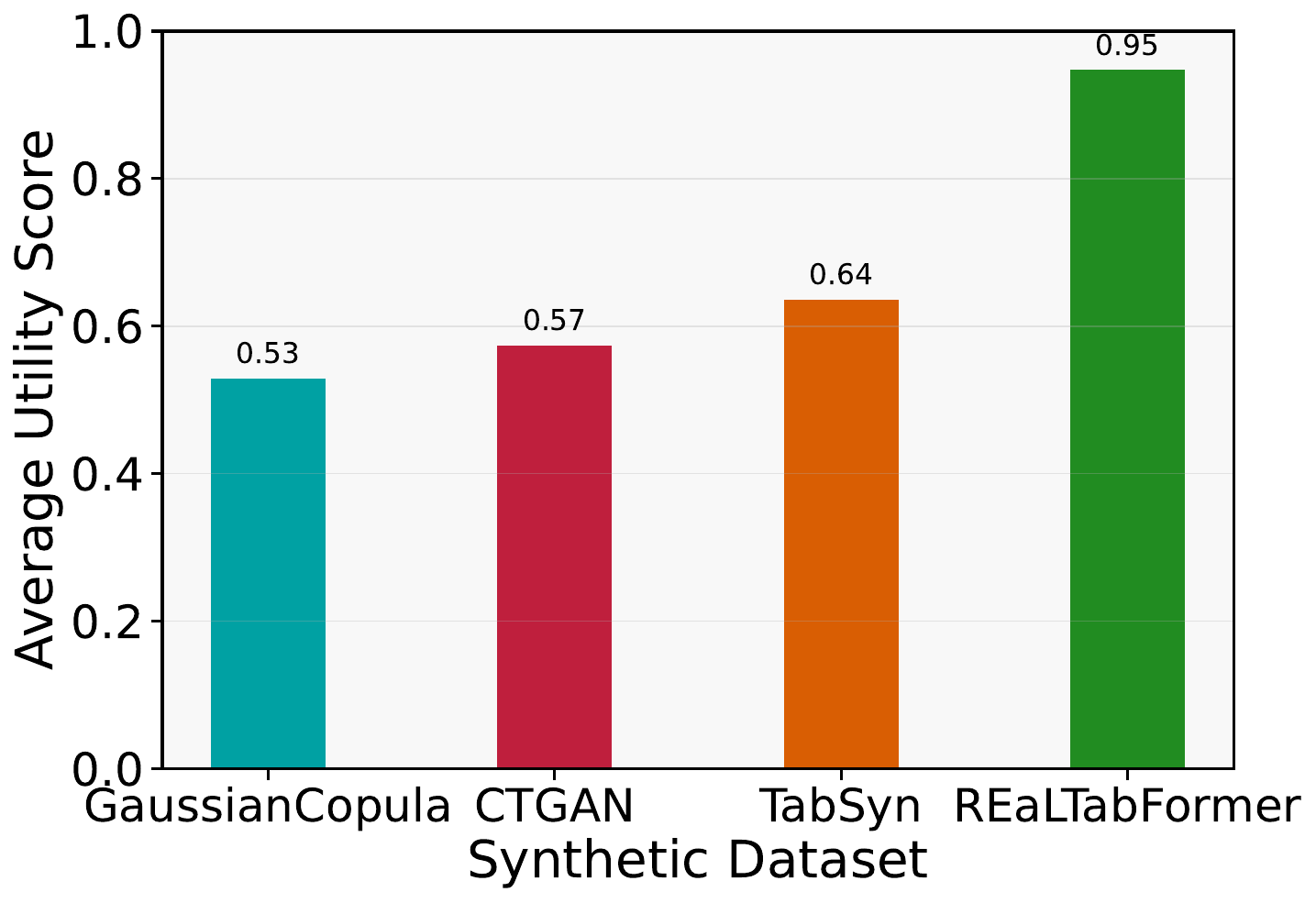}
    \caption{Average utility scores for pre-tactical arrival delay prediction across synthetic data generators.}
    \label{fig:arrival_pre_utility}
\end{figure}

Despite the increased task complexity, REaLTabFormer maintained  utility performance at 0.95 (Figure~\ref{fig:arrival_pre_utility}), demonstrating robustness across different prediction scenarios. This consistency suggests that the transformer-based approach captures the underlying statistical structure that relates scheduled flight information to eventual arrival performance, even when that relationship becomes increasingly attenuated by intervening factors.

\begin{figure}[htbp]
    \centering
    \includegraphics[width=\linewidth]{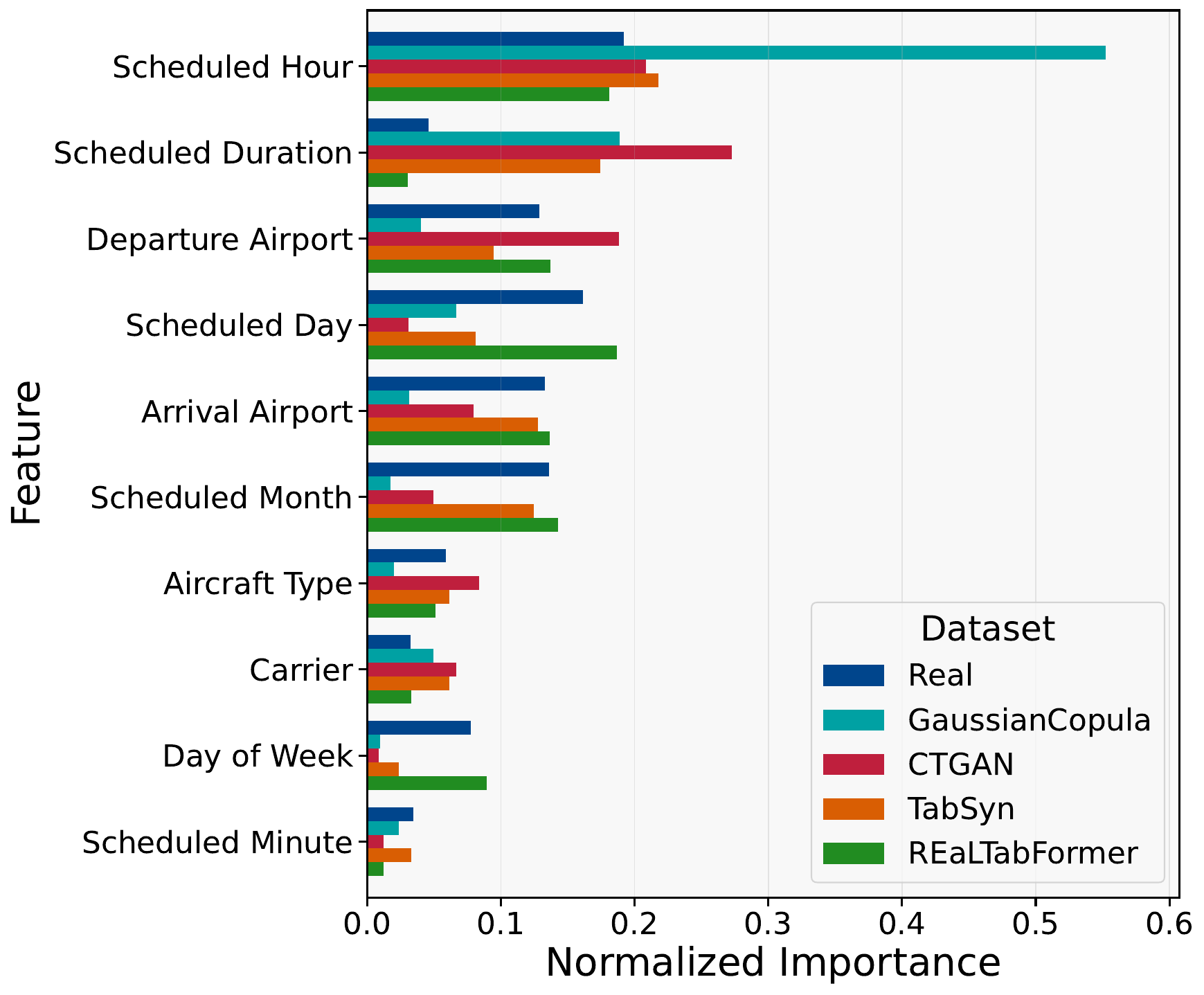}
    \caption{Feature importance comparison for pre-tactical arrival delay prediction averaged across all models.}
    \label{fig:arrival_features}
\end{figure}

Feature importance analysis for arrival delays revealed a slightly different pattern than departure delay prediction (Figure~\ref{fig:arrival_features}). Scheduled flight duration emerged as an additional significant predictor alongside temporal and airport features, reflecting the reality that longer flights have more opportunities for en-route delays and recovery. The distribution of importance across multiple features indicates that arrival delay prediction requires modeling interactions between temporal, spatial, and operational characteristics simultaneously.

\begin{figure}[htbp]
    \centering
    \includegraphics[width=0.8\linewidth]{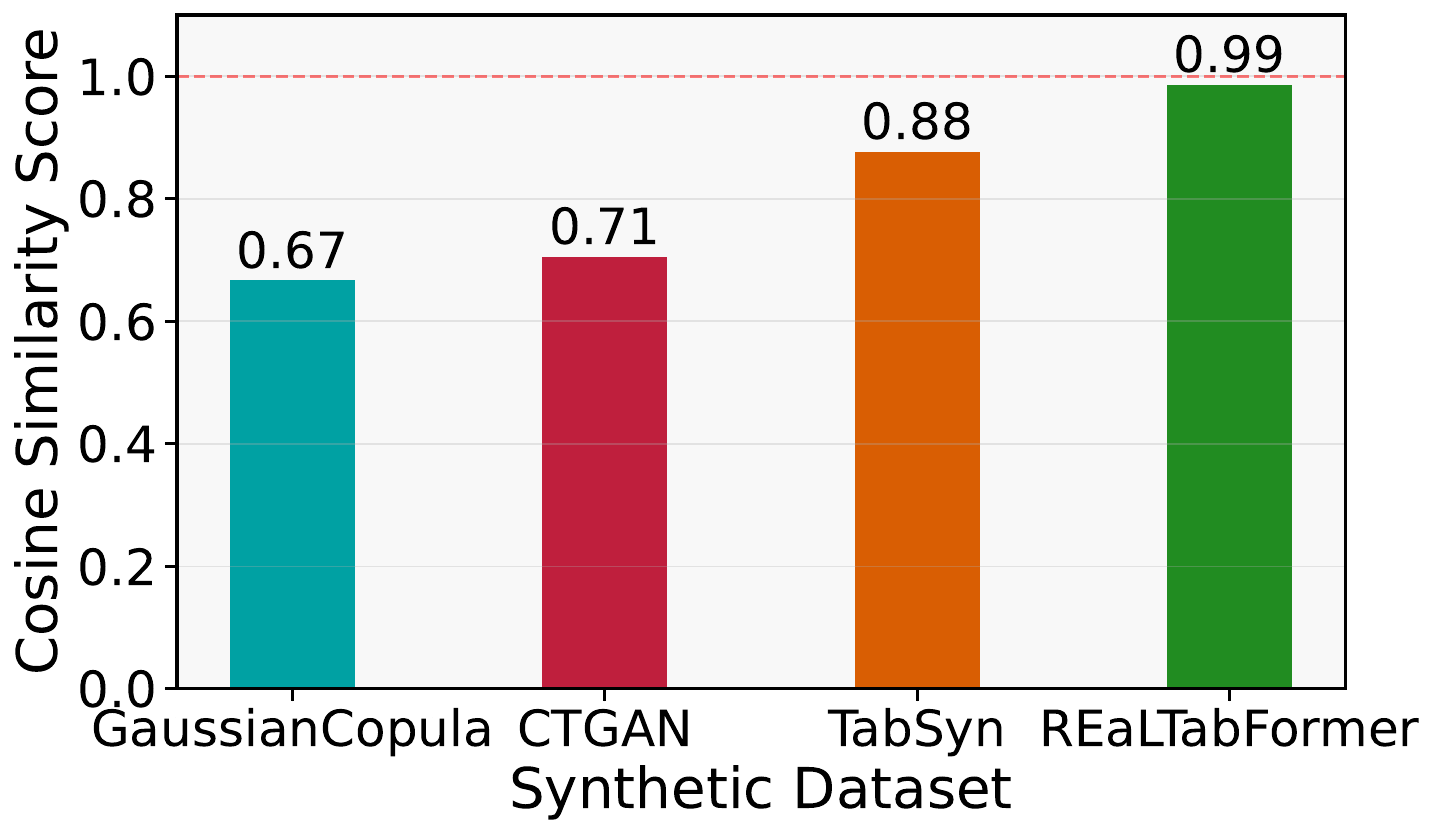}
    \caption{Average feature importance alignment scores for pre-tactical arrival delay prediction.}
    \label{fig:arrival_alignment}
\end{figure}

The feature alignment scores maintained the established pattern, with REaLTabFormer achieving near-perfect alignment (0.99) while simpler methods showed progressive degradation (Figure~\ref{fig:arrival_alignment}). This consistency in feature relationship preservation across increasingly difficult tasks demonstrates REaLTabFormer's fundamental advantage in capturing the complex dependencies present in aviation operational data.

\subsection{Turnaround Time Prediction}

\begin{figure}[htbp]
    \centering
    \includegraphics[width=\linewidth]{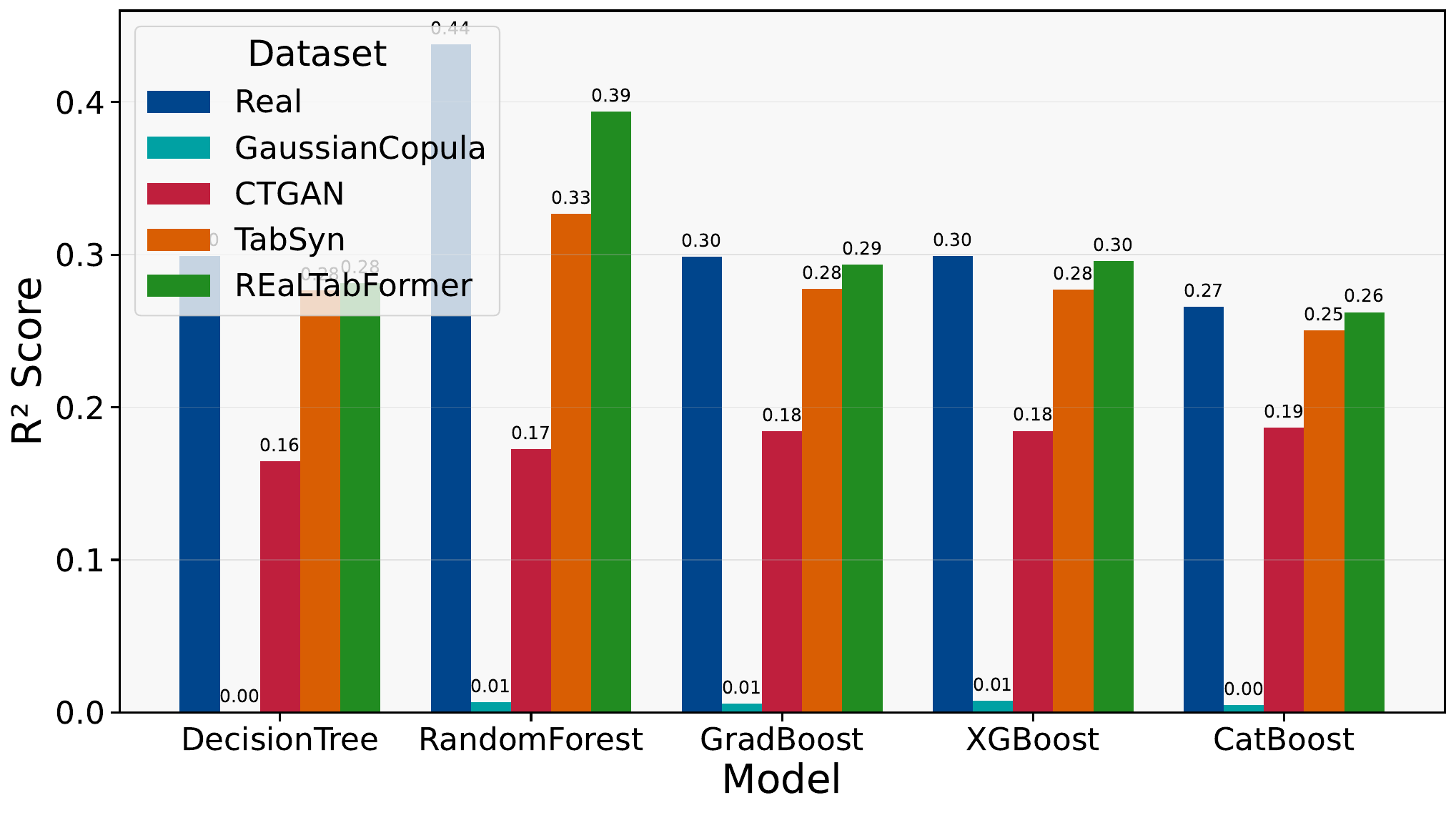}
    \caption{Coefficient of determination (R²) for pre-tactical turnaround time prediction across different models and synthetic data generators.}
    \label{fig:turnaround_pre_r2}
\end{figure}

Turnaround time prediction in pre-tactical mode demonstrated the highest predictability among all evaluated tasks, with real-data models achieving R² values between 0.27-0.44 (Figure~\ref{fig:turnaround_pre_r2}). This performance likely reflects the more deterministic nature of turnaround processes compared to delay propagation. While external factors certainly influence turnaround times, the core processes—passenger deplaning, cleaning, catering, fueling, and boarding—follow more predictable patterns based on aircraft type, airport facilities, and scheduled timing.

\begin{figure}[htbp]
    \centering
    \begin{subfigure}[b]{0.49\textwidth}
        \includegraphics[width=\linewidth]{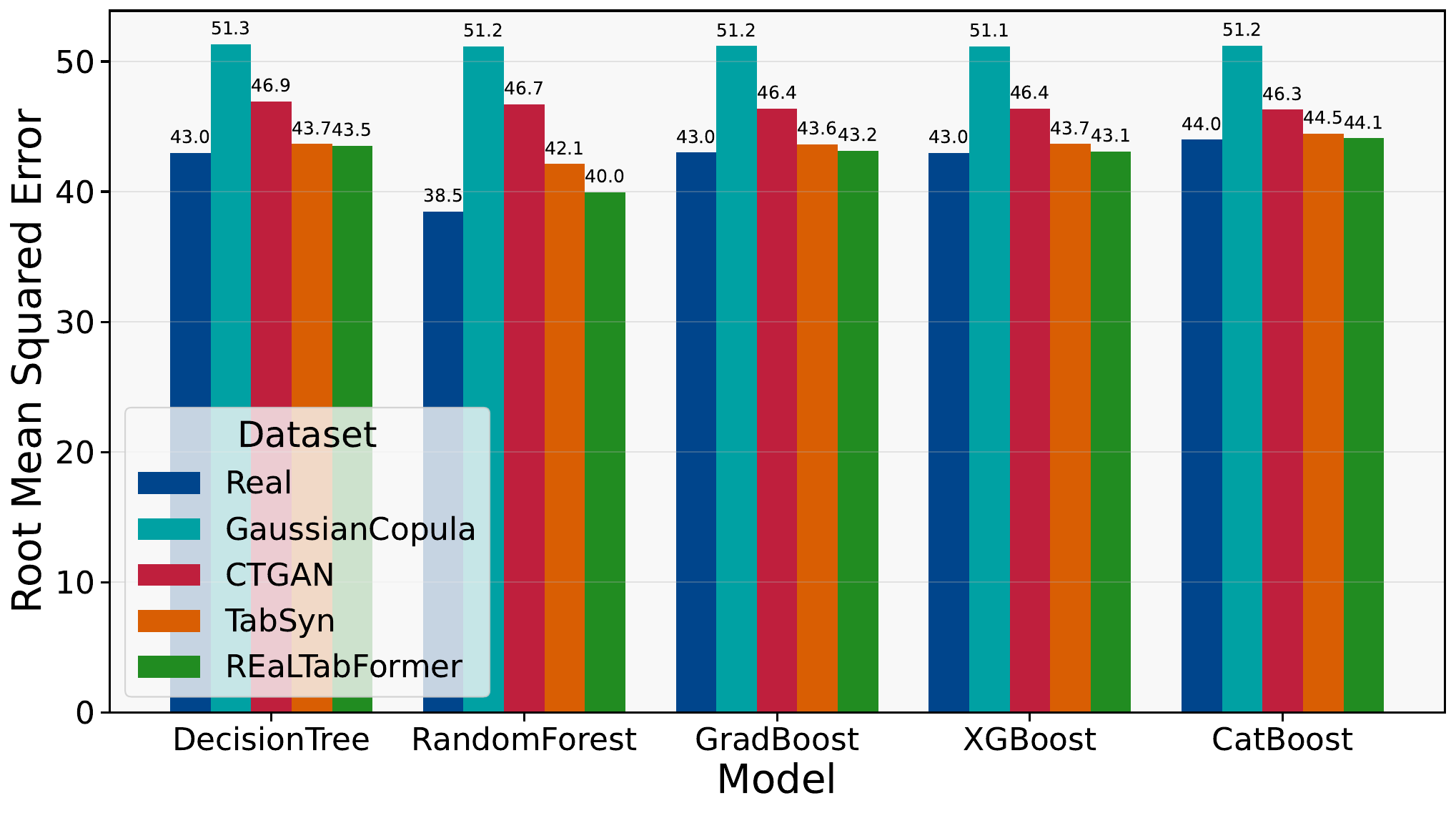}
        \caption{Root Mean Squared Error}
        \label{fig:turnaround_pre_rmse}
    \end{subfigure}
    \hfill
    \begin{subfigure}[b]{0.49\textwidth}
        \includegraphics[width=\linewidth]{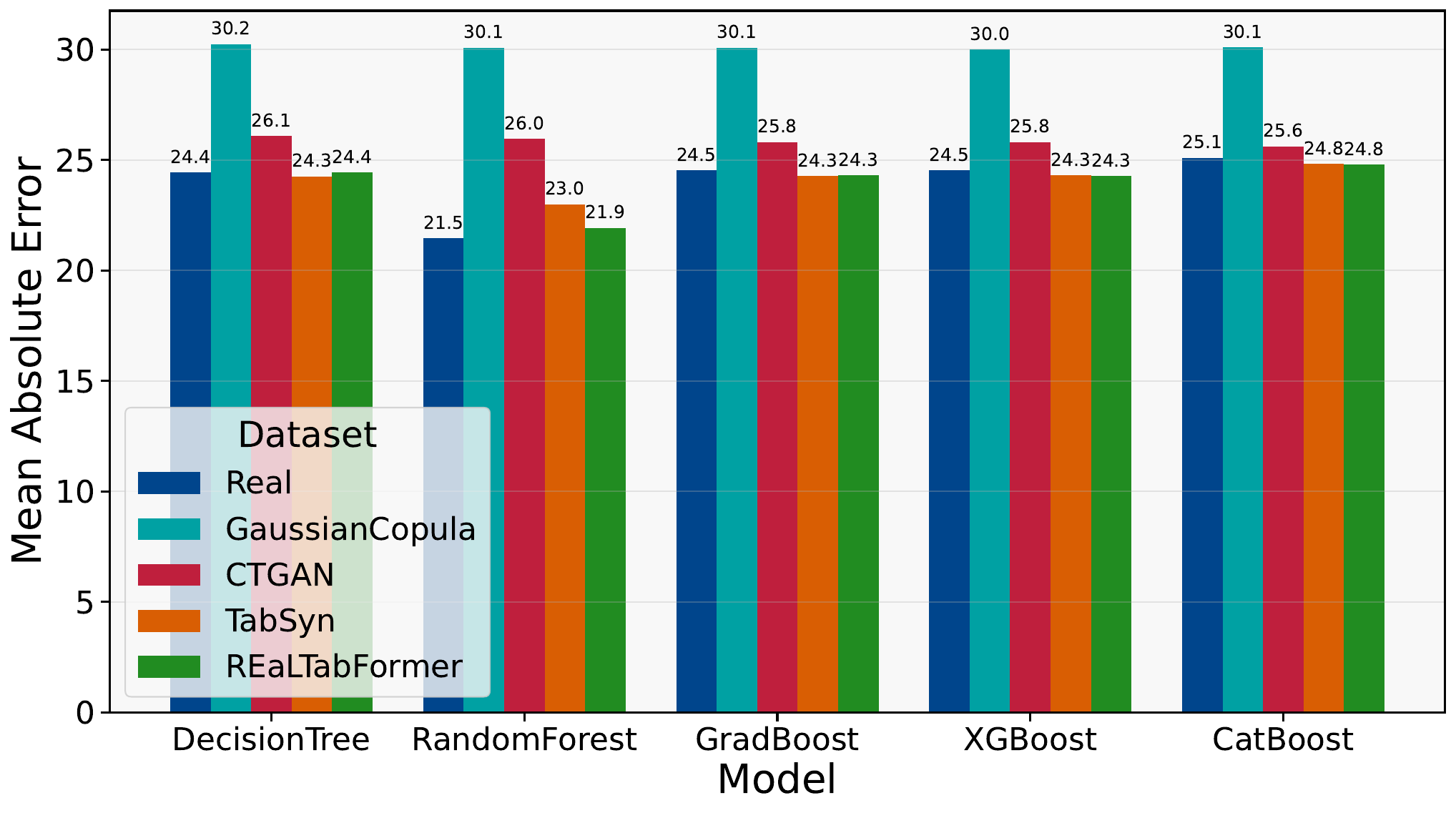}
        \caption{Mean Absolute Error}
        \label{fig:turnaround_pre_mae}
    \end{subfigure}
    \caption{Prediction error metrics for pre-tactical turnaround time across models and synthetic data generators.}
\end{figure}

The error metrics revealed pronounced performance differences between synthetic generators (Figures~\ref{fig:turnaround_pre_rmse} and \ref{fig:turnaround_pre_mae}). REaLTabFormer achieved RMSE values within 3\% of real-data baselines, demonstrating preservation of the predictive relationships between scheduled characteristics and turnaround requirements. TabSyn followed closely, while CTGAN showed moderate degradation and Gaussian Copula exhibited substantial performance gaps.

\begin{figure}[htbp]
    \centering
    \includegraphics[width=0.8\linewidth]{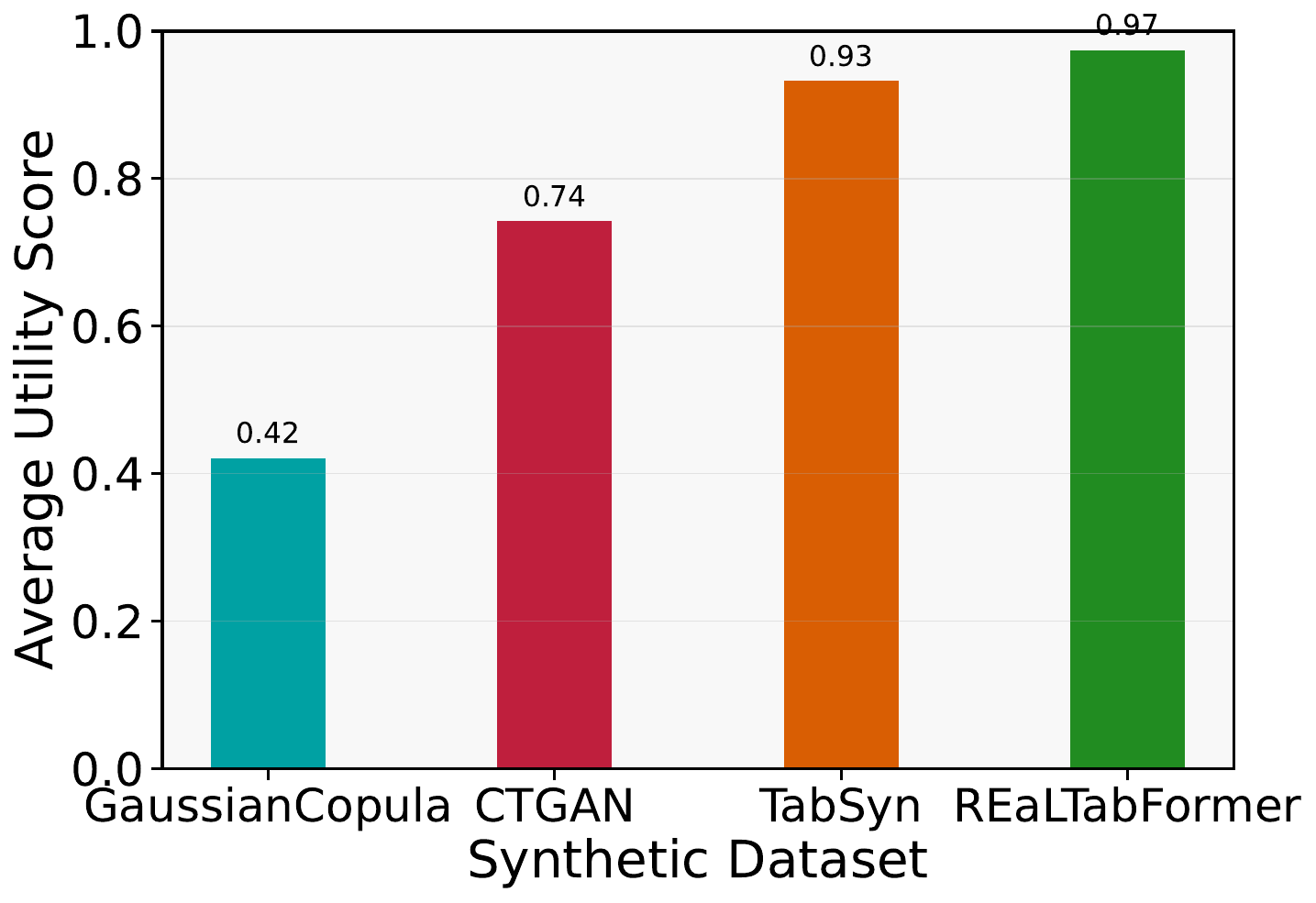}
    \caption{Average utility scores for pre-tactical turnaround time prediction across synthetic data generators.}
    \label{fig:turnaround_pre_utility}
\end{figure}

The utility score analysis confirmed turnaround prediction as the most successful application of synthetic data (Figure~\ref{fig:turnaround_pre_utility}). REaLTabFormer achieved a utility score of 0.97, followed by TabSyn at 0.93. These high scores indicate that for the operational task of turnaround planning, synthetic data can serve as an effective substitute for proprietary operational records, enabling advanced analytics capabilities for organizations without access to comprehensive historical datasets.

\begin{figure}[htbp]
    \centering
    \includegraphics[width=\linewidth]{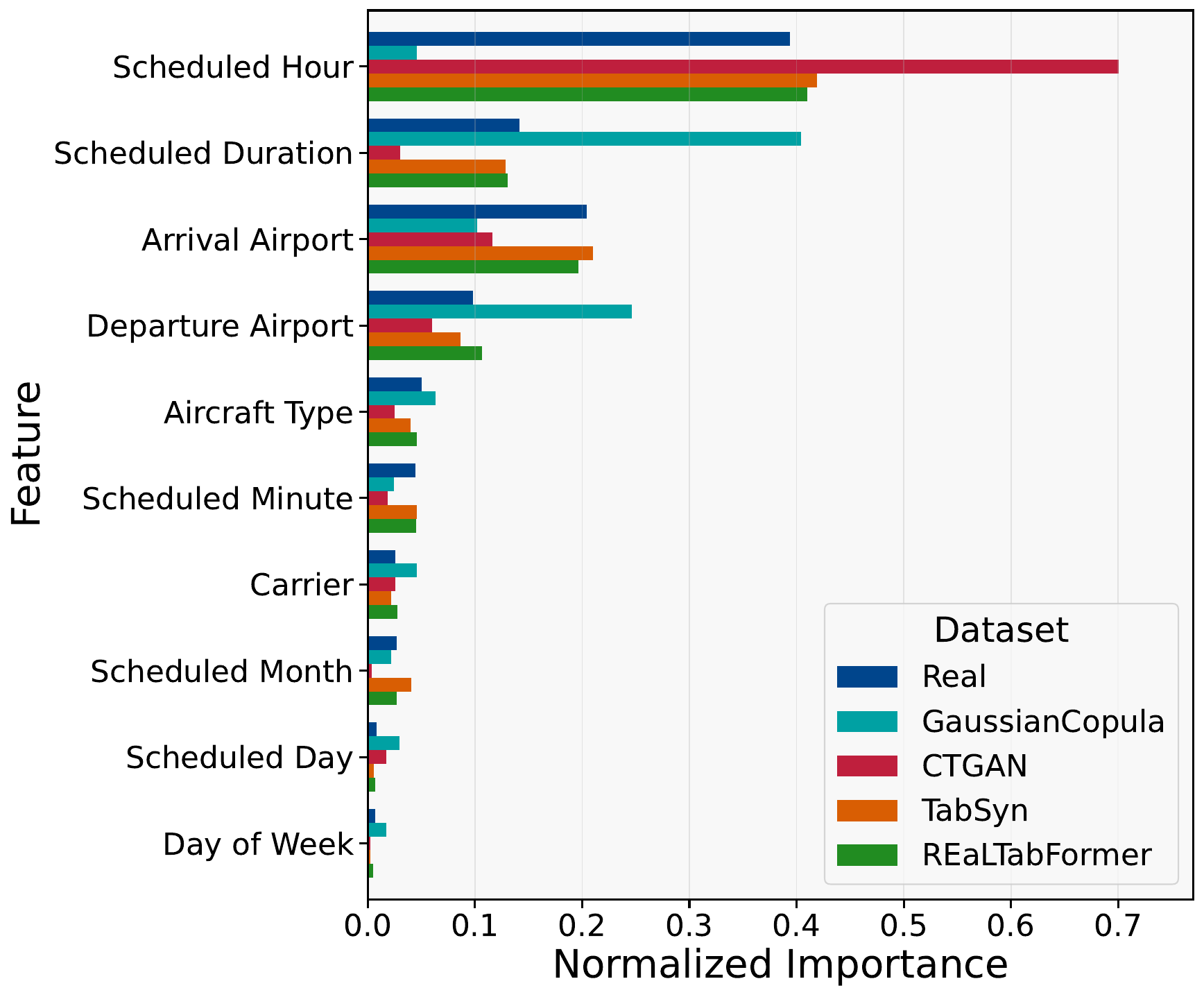}
    \caption{Feature importance comparison for pre-tactical turnaround time prediction averaged across all models.}
    \label{fig:turnaround_pre_features}
\end{figure}

Feature importance analysis revealed that scheduled hour/Duration and airport identifiers as the dominant predictors across all data sources (Figure~\ref{fig:turnaround_pre_features}). This pattern aligns with operational knowledge: turnaround processes exhibit significant time-of-day variations due to staffing levels, gate availability, and passenger flow patterns. Airport-specific effects capture differences in terminal layouts, ground handling procedures, and infrastructure constraints that systematically influence turnaround efficiency.

\begin{figure}[htbp]
    \centering
    \includegraphics[width=0.8\linewidth]{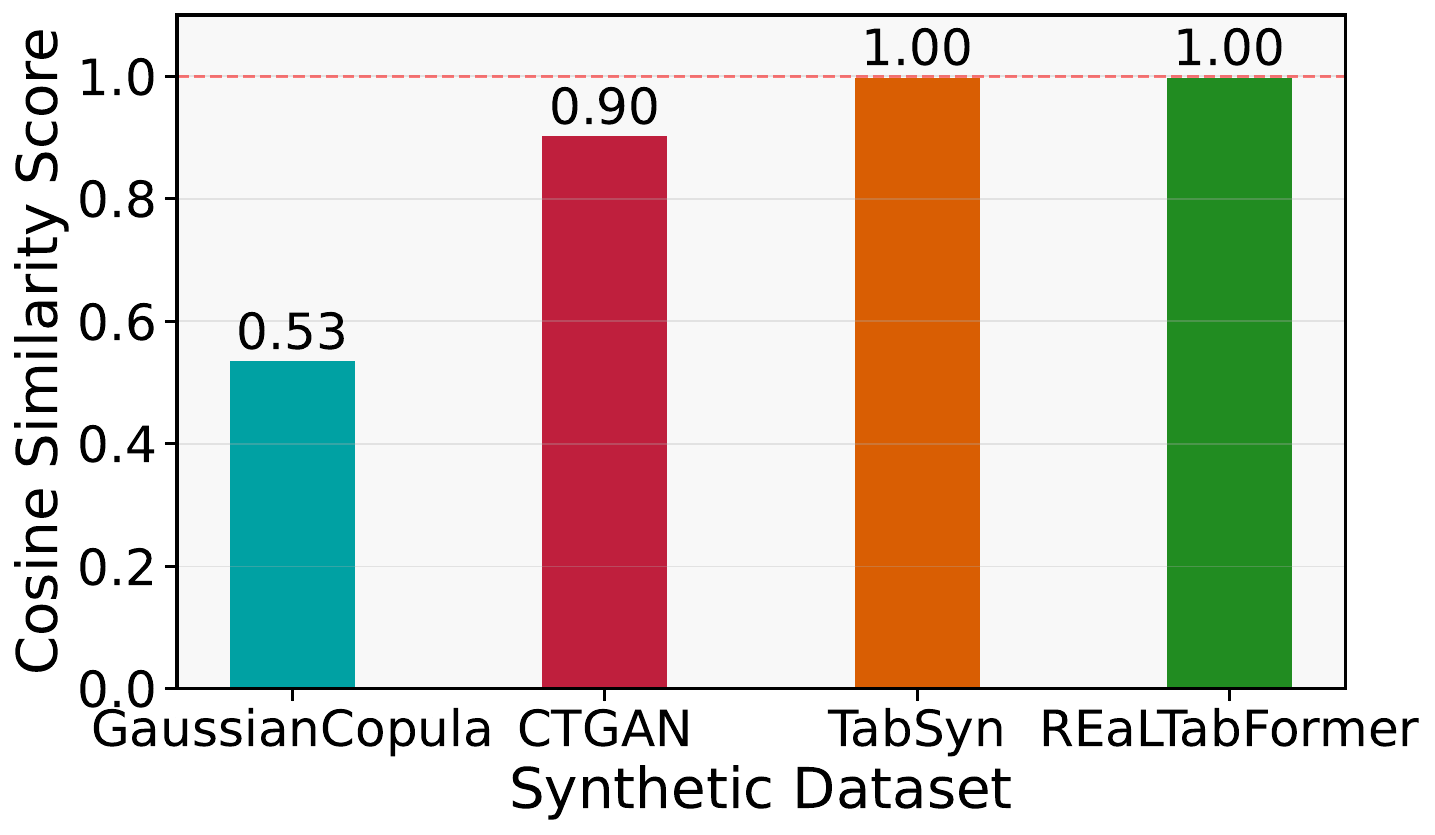}
    \caption{Average feature importance alignment scores for pre-tactical turnaround time prediction.}
    \label{fig:turnaround_pre_alignment}
\end{figure}

The feature importance alignment (Figure~\ref{fig:turnaround_pre_alignment}) shows that  REaLTabFormer and TabSyn achieved perfect alignment scores (1.00), ensuring that models trained on their synthetic data identify the same operational drivers as real-data models. This preservation of causal relationships is crucial for operational applications where understanding which factors drive turnaround variations is as important as prediction accuracy itself.

\subsection{Cross-Task Performance Analysis}

Analyzing results across all pre-tactical prediction tasks reveals several patterns that inform both synthetic data selection and operational expectations:

\textbf{Consistent Generator Hierarchy:} Across all tasks, the performance ranking remained stable: REaLTabFormer (95-97\% utility) $>$ TabSyn (64-93\%) $>$ CTGAN (60-74\%) $>$ Gaussian Copula (42-56\%). This consistency suggests fundamental differences in how each approach captures the relationships present in aviation operational data.

\textbf{Feature Relationship Preservation:} Advanced generators (REaLTabFormer, TabSyn) consistently preserved feature importance patterns across all tasks, while simpler methods showed task-dependent degradation. This preservation is crucial for operational deployment, where understanding causal relationships guides resource allocation and process improvement decisions.

\textbf{Fundamental Predictability Limits:} Even with real data, pre-tactical R² values peaked at 0.44 (turnaround), 0.36 (departure delay), and 0.34 (arrival delay). These ceilings reflect inherent uncertainty in aviation operations and should inform realistic expectations for any predictive system, regardless of data source.

\subsection{Operational Implications}

These findings carry important implications for aviation stakeholders:

\textbf{Strategic Planning Viability:} The high utility scores (95-97\%) achieved by REaLTabFormer indicate that synthetic data can effectively support critical pre-tactical decisions including crew scheduling, gate planning, and resource allocation made hours to days in advance. This capability could democratize advanced analytics across the industry.

\textbf{Research and Development Acceleration:} Preserved feature relationships make synthetic data valuable for algorithm development and benchmarking, allowing researchers to prototype pre-tactical prediction methods without accessing proprietary operational schedules. This could accelerate innovation in aviation analytics.

\textbf{Realistic Expectation Setting:} The modest R² values achieved even with real data highlight inherent uncertainties in pre-tactical prediction. Organizations should design planning processes that accommodate this uncertainty through robust buffers and contingency mechanisms rather than expecting precise predictions.

\textbf{Generator Selection Guidelines:} 
\begin{itemize}
    \item Use REaLTabFormer when prediction accuracy and feature relationship preservation are critical
    \item Consider TabSyn for applications requiring good performance with faster training times
    \item Reserve CTGAN for rapid prototyping scenarios where moderate accuracy loss is acceptable
    \item Limit Gaussian Copula to simple analyses focused primarily on marginal distributions
\end{itemize}

\section{Conclusion}

This study evaluated synthetic flight data as an alternative to real operational data for pre-tactical aviation predictions. Among four generators tested, REaLTabFormer achieved 95-97\% of real-data performance across turnaround time, departure delay, and arrival delay predictions while preserving critical feature relationships. Additionally, comprehensive fidelity assessments confirmed that synthetic data maintained essential statistical properties—including distributional similarity, correlation structures, and joint dependencies—before deployment in prediction tasks.
The analysis revealed fundamental predictability limits in pre-tactical scenarios—R² values peaked at 0.44 even with real data, reflecting inherent operational uncertainties when only scheduled information is available. This establishes realistic baselines for any predictive system regardless of data source.
These findings indicate that high-fidelity synthetic data can enable organizations without access to proprietary records to develop pre-tactical prediction capabilities. The consistent performance hierarchy (REaLTabFormer $>$ TabSyn $>$ CTGAN $>$ Gaussian Copula) provides guidance for method selection based on accuracy requirements. However, stakeholders should design planning processes that accommodate prediction uncertainty through appropriate buffers rather than expecting precise forecasts, given the stochastic nature of aviation operations.

\section*{Acknowledgment}

This paper is based on research conducted within the SynthAIr project, which has received funding from the SESAR Joint Undertaking under the European Union’s Horizon Europe research and innovation program (grant agreement No. 101114847). The views and opinions expressed in this paper are solely those of the authors and do not necessarily reflect those of the European Union or the SESAR 3 Joint Undertaking. Neither the European Union nor the SESAR 3 Joint Undertaking can be held responsible for any use of the information contained herein.

\balance

\bibliographystyle{IEEEtran}
\bibliography{references}

\appendices
\section{Generative Models Overview}
\label{sec:appendix-generative_models}

\subsection{Gaussian Copula (GC)}
A statistical approach that models multivariate dependencies by separating marginal distributions from their dependence structure. This method is particularly suited for mixed-type tabular data, handling both categorical (airports, carriers, aircraft types) and continuous (delays, turnaround times) variables through a unified framework.

The Gaussian copula model operates in two stages:

\textbf{Marginal Distribution Fitting:} For each variable $X_i$ in the dataset, we fit an appropriate univariate distribution. For continuous variables, the implementation employs a flexible selection mechanism that evaluates multiple candidate distributions (e.g., Gaussian, exponential, beta) using the Kolmogorov-Smirnov test, automatically selecting the best-fitting model. Categorical variables are encoded using ordinal encoding, preserving their discrete nature while enabling numerical processing.

\textbf{Dependency Modeling:} After fitting marginals, we transform each variable to the standard normal domain using:
\begin{equation}
Z_i = \Phi^{-1}(F_i(X_i))
\end{equation}
where $F_i$ is the fitted CDF for variable $i$ and $\Phi^{-1}$ is the inverse standard normal CDF. The transformation clips extreme values to $[\epsilon, 1-\epsilon]$ to ensure numerical stability.

The correlation structure is then captured by computing the correlation matrix $\mathbf{R}$ of the transformed variables $\{Z_1, ..., Z_d\}$. For random variables $X_1, X_2, \ldots, X_d$ with marginal CDFs $F_1, F_2, \ldots, F_d$, let $u_i = F_i(X_i)$ represent the probability integral transforms. The joint distribution is then modeled using the Gaussian copula:
\begin{equation}
C_{\mathbf{R}}(u_1, ..., u_d) = \Phi_{\mathbf{R}}(\Phi^{-1}(u_1), ..., \Phi^{-1}(u_d))
\end{equation}
where $\Phi_{\mathbf{R}}$ represents the CDF of a multivariate normal distribution with correlation matrix $\mathbf{R}$.

\textbf{Synthetic Data Generation:} To generate new samples:
\begin{enumerate}
    \item Sample $\mathbf{z} \sim \mathcal{N}(\mathbf{0}, \mathbf{R})$ from the multivariate normal
    \item Transform to uniform: $u_i = \Phi(z_i)$ for each component
    \item Apply inverse marginal CDFs: $x_i = F_i^{-1}(u_i)$
    \item For categorical variables, round to nearest valid category index and decode
\end{enumerate}

This approach preserves parametrically-fitted marginal distributions while capturing linear correlations between variables. However, it may struggle with complex non-linear dependencies and rare event combinations that are particularly important in aviation operations.

\subsection{Conditional Tabular GAN (CTGAN)}

CTGAN \cite{xu2019modeling} is a generative adversarial network specifically designed for mixed-type tabular data, addressing the unique challenges of generating realistic flight records with both continuous (delays, durations) and categorical (airports, carriers) variables.

\begin{figure*}[htbp]
    \centering
    \includegraphics[width=\linewidth]{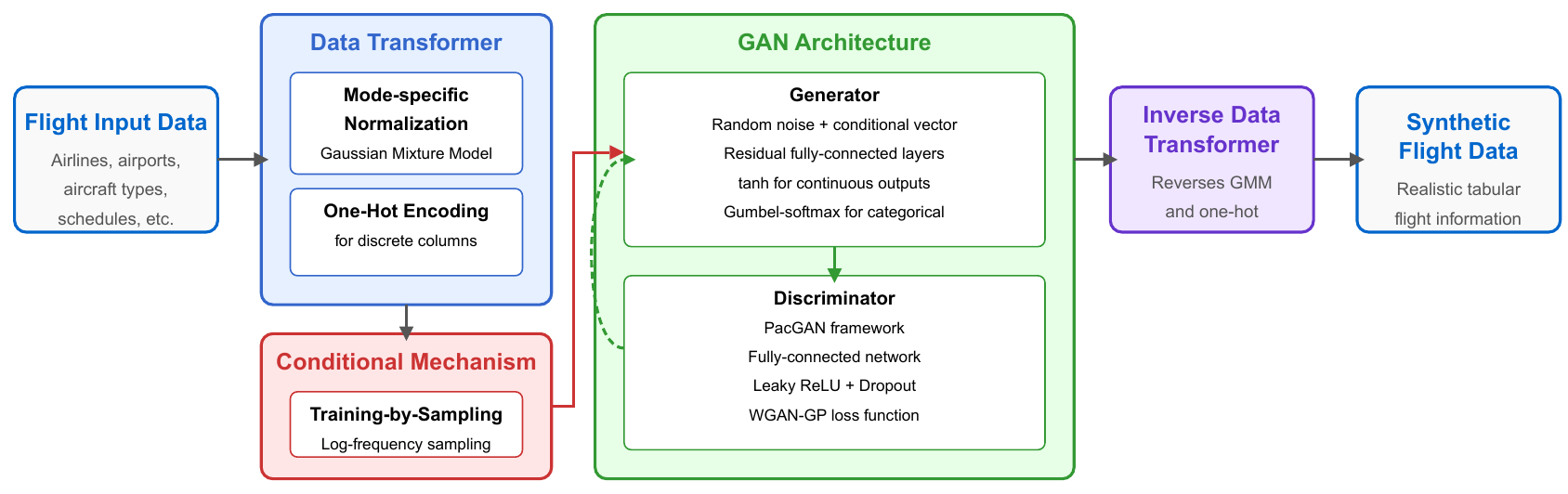}
    \caption{CTGAN Architecture for Flight Data Generation. The diagram illustrates the complete pipeline for synthetic flight data generation. Starting with flight input data (left), the system transforms it using mode-specific normalization for continuous variables and one-hot encoding for discrete columns like airline codes and airports. The conditional mechanism employs training-by-sampling with log-frequency to address category imbalance. The GAN architecture consists of a generator (combining random noise with conditional vectors) and discriminator (using PacGAN and WGAN-GP loss). Finally, the inverse data transformer converts the generator's output back into realistic flight data format, preserving the complex statistical relationships present in the original data.}
    \label{fig:ctgan_architecture}
\end{figure*}

\textbf{Architecture Overview:} As illustrated in Figure~\ref{fig:ctgan_architecture}, CTGAN consists of four interconnected components that transform raw flight data into a generative model capable of producing new synthetic samples:

\textbf{1. Data Transformer:} The first component in the pipeline (shown in blue in Figure~\ref{fig:ctgan_architecture}) preprocesses mixed-type data for neural network processing:
\begin{itemize}
    \item \textbf{Mode-specific Normalization}: For continuous variables, employs a Bayesian Gaussian Mixture Model (BGM) to identify multimodal distributions:
    \begin{itemize}
        \item Fits a BGM with up to 10 components using variational inference
        \item Each value is represented as: (1) normalized scalar within its mode, and (2) one-hot vector indicating the mode
        \item Prevents gradient vanishing in non-Gaussian distributions common in flight delays
    \end{itemize}
    \item \textbf{One-hot Encoding}: Categorical variables (airlines, airports, aircraft types) are encoded as one-hot vectors
\end{itemize}

\textbf{2. Conditional Mechanism:} The conditional component (shown in red in Figure~\ref{fig:ctgan_architecture}) addresses severe category imbalance in aviation data:
\begin{itemize}
    \item \textbf{Training-by-Sampling}: Uses log-frequency sampling to ensure rare categories receive adequate training:
    \begin{equation}
    P(c_i) = \frac{\log(f_i) + 1}{\sum_j (\log(f_j) + 1)}
    \end{equation}
    where $f_i$ is the frequency of category $i$
    \item \textbf{Conditional Vector}: Generator receives a one-hot vector indicating which discrete column/value to generate, enabling balanced learning across all categories
\end{itemize}

\textbf{3. GAN Architecture:} The core generative components (central green section in Figure~\ref{fig:ctgan_architecture}) consist of:

\textbf{Generator Network:} 
\begin{itemize}
    \item \textbf{Architecture}: Two residual blocks with batch normalization and ReLU activation
    \begin{itemize}
        \item Input: Random noise (128-dim) concatenated with conditional vector
        \item Hidden layers: 256 neurons each with residual connections
        \item Output: Mixed activations - tanh for normalized values, Gumbel-softmax for categories
    \end{itemize}
    \item \textbf{Residual Block Structure}:
    \begin{equation}
    \text{Residual}(x) = \text{Concat}[\text{ReLU}(\text{BN}(\text{FC}(x))), x]
    \end{equation}
\end{itemize}

\textbf{Discriminator Network:}
\begin{itemize}
    \item \textbf{PacGAN Framework}: Processes 10 samples jointly to prevent mode collapse
    \item \textbf{Architecture}: Two hidden layers (256 neurons) with LeakyReLU ($\alpha$=0.2) and dropout (0.5)
    \item \textbf{WGAN-GP Loss}: Wasserstein loss with gradient penalty:
    \begin{equation}
    \mathcal{L}_D = -(\mathbb{E}[D(\text{real})] - \mathbb{E}[D(\text{fake})]) + \lambda \cdot GP
    \end{equation}
    where the gradient penalty term:
    \begin{equation}
    GP = \mathbb{E}_{\hat{x}} \left[ (||\nabla_{\hat{x}} D(\hat{x})||_2 - 1)^2 \right]
    \end{equation}
    with $\hat{x}$ interpolated between real and fake samples, and $\lambda = 10$
\end{itemize}

\textbf{4. Inverse Data Transformer:} The final component (purple in Figure~\ref{fig:ctgan_architecture}) reverses the initial transformations to produce flight data in its original format.

\textbf{Training Methodology:} The training process (detailed in the bottom section of Figure~\ref{fig:ctgan_architecture}) follows these steps:
\begin{enumerate}
    \item \textbf{Data Preprocessing}: Transform data using mode-specific normalization and one-hot encoding
    \item \textbf{Adversarial Training Loop}:
    \begin{itemize}
        \item Sample conditional vectors based on log-frequency
        \item Select real data matching the condition
        \item Generate fake samples: $G(z, c)$ where $z \sim \mathcal{N}(0, I)$ and $c$ is the condition
        \item Update discriminator with WGAN-GP loss
        \item Update generator with combined loss:
        \begin{equation}
        \mathcal{L}_G = -\mathbb{E}[D(G(z,c))] + \mathcal{L}_{\text{cond}}
        \end{equation}
        where $\mathcal{L}_{\text{cond}}$ is cross-entropy between generated categories and condition
    \end{itemize}
\end{enumerate}

\textbf{Implementation Details:}
\begin{itemize}
    \item Optimizers: Adam with learning rate $2 \times 10^{-4}$, $\beta_1 = 0.5$, $\beta_2 = 0.9$
    \item Weight decay: $10^{-6}$ for regularization
    \item Batch size: 500 samples
    \item Training epochs: 300-500 (determined by loss convergence)
    \item Discriminator updates per generator update: 1
    \item Gumbel-softmax temperature: 0.2 for discrete outputs
\end{itemize}

\textbf{Generation Process:}
\begin{enumerate}
    \item For conditional generation: specify discrete column and value
    \item Create conditional vector from specified condition
    \item Sample Gaussian noise: $z \sim \mathcal{N}(0, I_{128})$
    \item Concatenate noise with conditional vector
    \item Pass through generator to produce normalized outputs
    \item Apply inverse data transformation to recover original data format
\end{enumerate}

This architecture enables CTGAN to effectively handle the mixed-type nature of flight data while addressing common challenges such as mode collapse and training instability through its specialized components shown in Figure~\ref{fig:ctgan_architecture}.

\subsection{TabSyn}

TabSyn \cite{zhang2024mixed} addresses the challenge of synthesizing mixed-type tabular data by decoupling representation learning from sample generation. Rather than applying diffusion directly to raw columns—an approach that struggles with complex inter-column dependencies and often requires thousands of sampling steps—TabSyn implements a two-stage pipeline: first mapping data to a continuous latent space via a Variational Autoencoder (VAE), then applying score-based diffusion in this regularized embedding space.

\begin{figure*}[htbp]
    \centering
    \includegraphics[width=\linewidth]{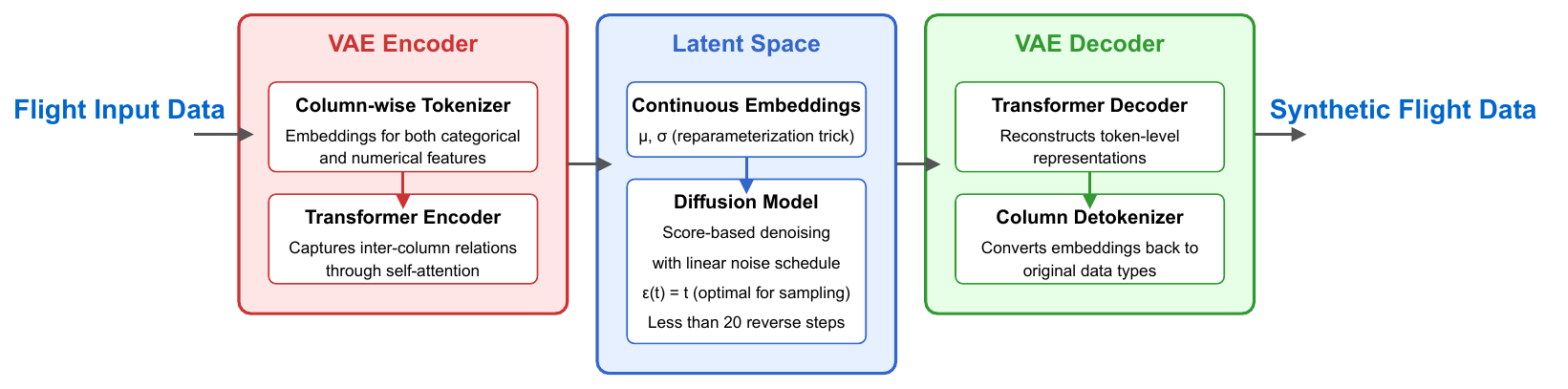}
    \caption{TabSyn architecture for flight data generation. The model employs a two-stage pipeline with three main components: (1) VAE Encoder (left, red) transforms mixed-type flight records into continuous latent vectors through column-wise tokenization and transformer-based encoding; (2) Latent Space Diffusion (center, blue) applies score-based denoising with a linear noise schedule, enabling efficient sampling in 20-50 steps; (3) VAE Decoder (right, green) reconstructs synthetic flight records through transformer decoding and column-specific detokenization.}
    \label{fig:tabsyn_architecture}
\end{figure*}

\textbf{Architecture Overview:} As illustrated in Figure~\ref{fig:tabsyn_architecture}, TabSyn consists of three integrated components that work together to transform raw flight data into high-quality synthetic records:

\textbf{1. VAE Encoder Module:} The encoder (shown in red in Figure~\ref{fig:tabsyn_architecture}) transforms raw flight data into continuous latent representations through two stages:
\begin{itemize}
    \item \textbf{Column-wise Tokenizer}: Processes each column independently based on its data type. For numerical attributes (e.g., flight duration, delays), a linear transformation maps values to token embeddings. For categorical attributes (e.g., aircraft type, airports), learned embedding lookup tables transform each category into dense vectors. The combined token dimension is set to 8 for aviation datasets.
    
    \item \textbf{Transformer Encoder}: After tokenization, embeddings are processed through a 2-layer transformer with single-head attention. This architecture captures non-linear relationships between flight attributes through self-attention mechanisms with a hidden dimension expansion factor of 32. The encoder outputs distribution parameters $(\mu, \sigma)$ for each input record, from which latent vectors are sampled using the reparameterization trick.
\end{itemize}

\textbf{2. Latent Space Diffusion Module:} The diffusion component (blue section in Figure~\ref{fig:tabsyn_architecture}) operates entirely in the continuous latent space:
\begin{itemize}
    \item \textbf{Forward Process}: Gradually corrupts latent vectors by adding Gaussian noise following a linear schedule, which has been shown to be optimal for this latent space formulation.
    
    \item \textbf{Denoising Network}: A 3-layer MLP with SiLU activations and 1024 hidden dimensions learns to reverse the diffusion process. The network incorporates positional embeddings to encode the noise level at each step. Training minimizes the Elucidating Diffusion Model (EDM) loss:
    \begin{equation}
    \mathcal{L}_{\text{EDM}} = \mathbb{E}_{z_0, t, \epsilon} \left[ \lambda(t) \|s_\theta(z_t, t) - \epsilon\|_2^2 \right]
    \end{equation}
    where $z_t$ is the noisy latent at time $t$, $s_\theta$ is the denoising network, and $\lambda(t)$ is a weighting function derived from the noise schedule.
    
    \item \textbf{Efficient Sampling}: The linear noise schedule enables generation with only 20-50 reverse diffusion steps, compared to thousands required by traditional diffusion approaches operating in data space.
\end{itemize}

\textbf{3. VAE Decoder Module:} The decoder (green section in Figure~\ref{fig:tabsyn_architecture}) mirrors the encoder structure to reconstruct tabular data:
\begin{itemize}
    \item \textbf{Transformer Decoder}: Processes sampled latent vectors through 2 transformer layers with identical configuration to the encoder, maintaining learned interdependencies while reconstructing token-level representations.
    
    \item \textbf{Column Detokenizer}: Maps token representations back to original data types using linear projections for numerical attributes and softmax layers for categorical attributes, ensuring type-appropriate reconstruction.
\end{itemize}

\textbf{Training Methodology:} TabSyn employs a sequential two-phase training process (bottom section of Figure~\ref{fig:tabsyn_architecture}):

\textbf{Phase 1 - VAE Training:}
The VAE is trained end-to-end with an adaptive KL-divergence weighting:
\begin{equation}
\mathcal{L}_{\text{VAE}} = \mathcal{L}_{\text{recon}} + \beta(t) \cdot \mathcal{L}_{\text{KL}}
\end{equation}
where:
\begin{itemize}
    \item Reconstruction loss combines MSE for numerical features and cross-entropy for categorical features
    \item $\beta(t)$ follows an adaptive schedule: initialized at $10^{-2}$, decaying to $10^{-5}$ with factor 0.7 every 5 epochs if above minimum
    \item Training runs for 200 epochs with Adam optimizer (learning rate $10^{-3}$) and batch size 8192
\end{itemize}

\textbf{Phase 2 - Diffusion Model Training:}
\begin{itemize}
    \item The pre-trained VAE encoder (with frozen weights) generates latent embeddings for all training data
    \item Embeddings are normalized by subtracting mean and scaling by factor 2
    \item The MLP diffusion model trains for 1000 epochs with learning rate $3 \times 10^{-4}$
    \item Early stopping with patience of 500 epochs based on validation loss
\end{itemize}

\textbf{Generation Process:}
\begin{enumerate}
    \item Sample initial noise: $z_T \sim \mathcal{N}(0, I)$ matching VAE latent dimensions
    \item Apply discretized reverse SDE for $T$ steps (typically 50):
    \begin{equation}
    z_{t-1} = z_t - \sigma_t(\sigma_t - \sigma_{t-1}) \cdot s_\theta(z_t, \sigma_t)
    \end{equation}
    \item Scale and shift final $z_0$ to match training embedding distribution
    \item Pass through VAE decoder to obtain synthetic tabular records
\end{enumerate}

This architecture enables TabSyn to capture complex dependencies in mixed-type aviation data while maintaining computational efficiency through its latent space formulation, making it practical for generating large-scale synthetic flight datasets.

\subsection{REaLTabFormer}

REaLTabFormer \cite{solatorio2023realtabformer} leverages transformer architectures originally designed for natural language processing to generate synthetic tabular data. Unlike traditional tabular generators that process columns independently, REaLTabFormer treats each flight record as a sequence of tokens, enabling the model to capture complex dependencies between features through autoregressive generation.

\begin{figure*}[htbp]
    \centering
    \includegraphics[width=\linewidth]{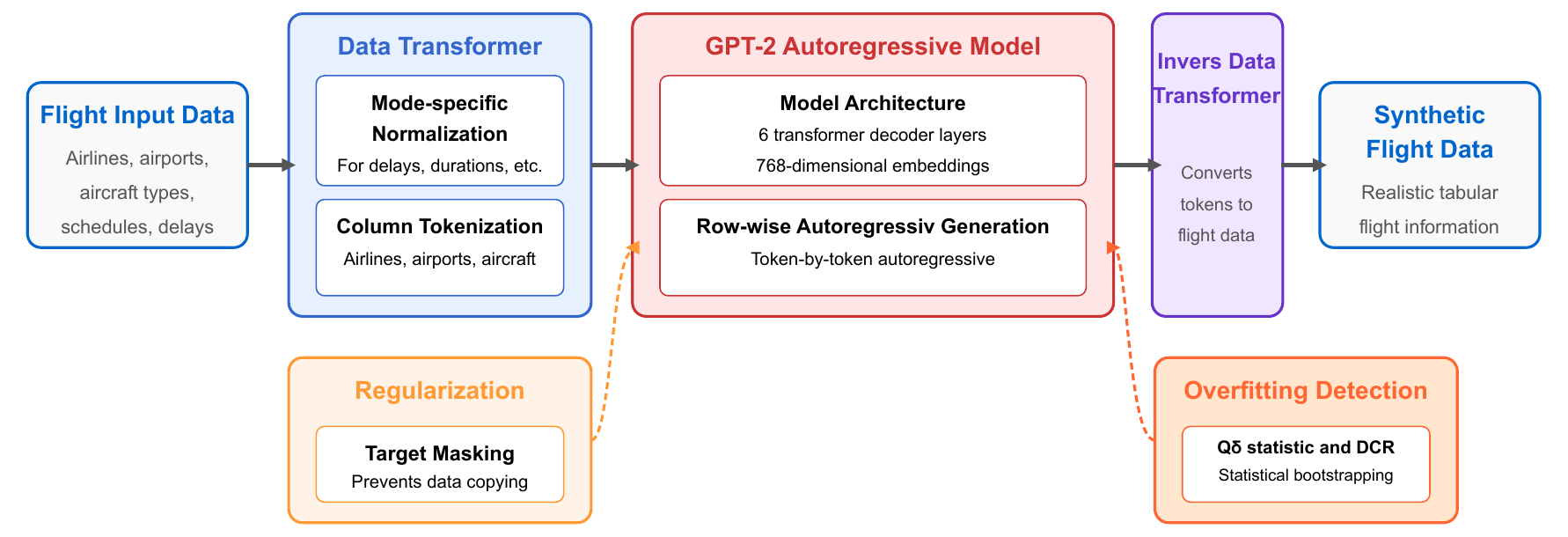}
    \caption{REaLTabFormer architecture for tabular flight data generation. The pipeline consists of four main components: (1) Data Transformer applies mode-specific normalization for continuous variables and column tokenization for categorical fields; (2) GPT-2 Autoregressive Model with 6 transformer decoder layers generates flight records row-by-row, token-by-token; (3) Regularization component implements target masking to prevent memorization; (4) Overfitting Detection uses Q$_\delta$ statistic and Distance to Closest Record (DCR) measurements to ensure model quality. The Inverse Data Transformer converts generated tokens back into structured flight data format.}
    \label{fig:rtf_architecture}
\end{figure*}

\textbf{Architecture Overview:} As illustrated in Figure~\ref{fig:rtf_architecture}, REaLTabFormer consists of four interconnected components that transform flight data into token sequences and generate new records through autoregressive prediction:

\textbf{1. Data Transformer:} The data transformer (blue section in Figure~\ref{fig:rtf_architecture}) preprocesses mixed-type flight data into tokenized representations:
\begin{itemize}
    \item \textbf{Mode-specific Normalization}: For continuous variables (delays, durations, scheduled times):
    \begin{itemize}
        \item Applies text-based representation strategies tailored to each data type
        \item Numeric values undergo precision-controlled string conversion and are split into fixed-length substrings (default single-character tokens)
        \item Datetime columns are converted to Unix timestamps, shifted by column mean to reduce scale, then processed through the numeric pipeline
    \end{itemize}
    \item \textbf{Column Tokenization}: For categorical variables (airlines, airports, aircraft types):
    \begin{itemize}
        \item Creates column-specific vocabulary mappings
        \item Preserves unique value distributions within each categorical column
        \item Enables the model to learn column-specific patterns and constraints
    \end{itemize}
\end{itemize}

\textbf{2. GPT-2 Autoregressive Model:} The core generative component (red section in Figure~\ref{fig:rtf_architecture}) uses a GPT-2 architecture optimized for tabular data:
\begin{itemize}
    \item \textbf{Model Configuration}: 
    \begin{itemize}
        \item 6 transformer decoder layers (DistilGPT-2 scale)
        \item 768-dimensional embeddings with 12 attention heads
        \item Special token handling for sequence boundaries ([BOS], [EOS]) and missing values
        \item Maximum position embeddings adjusted to accommodate the longest tokenized records
    \end{itemize}
    \item \textbf{Row-wise Generation}: Each flight record is treated as a sequence where:
    \begin{itemize}
        \item Tokens are generated left-to-right, conditioned on all previous tokens
        \item The model learns to predict $P(x_i | x_1, ..., x_{i-1})$ for each token position
        \item This autoregressive approach naturally captures dependencies between features
    \end{itemize}
\end{itemize}

\textbf{3. Regularization Mechanism:} The regularization component (orange section in Figure~\ref{fig:rtf_architecture}) prevents overfitting and memorization:
\begin{itemize}
    \item \textbf{Target Masking}: Randomly replaces 10\% of tokens with [RMASK] tokens during training
    \item Forces the model to learn contextual relationships rather than memorizing exact patterns
    \item The mask rate parameter controls regularization strength
\end{itemize}

\textbf{4. Overfitting Detection:} The monitoring system (bottom orange section in Figure~\ref{fig:rtf_architecture}) ensures generation quality:
\begin{itemize}
    \item \textbf{Q$_\delta$ Statistic}: Measures dissimilarity between synthetic and real data distributions:
    \begin{equation}
    Q_\delta = \frac{1}{N}\sum_{q} |p_q - q|
    \end{equation}
    where $q$ is a quantile in the evaluation set and $p_q$ is the proportion of synthetic samples below the value at quantile $q$
    \item \textbf{Distance to Closest Record (DCR)}: Evaluates whether synthetic samples are copies of training data
    \item \textbf{Statistical Bootstrapping}: Establishes data-specific thresholds through 500 bootstrap rounds
    \item Training stops when the statistic breaches its threshold in consecutive checks
\end{itemize}

\textbf{Training Methodology:}
\begin{enumerate}
    \item \textbf{Data Preprocessing}: Transform flight data using mode-specific strategies for continuous variables and tokenization for categorical columns
    \item \textbf{Vocabulary Construction}: Build column-specific vocabularies with special tokens
    \item \textbf{Autoregressive Training}: Train the model to predict next token given previous tokens:
    \begin{equation}
    \mathcal{L} = -\sum_{i=1}^{n} \log P(x_i | x_1, ..., x_{i-1})
    \end{equation}
    \item \textbf{Overfitting Monitoring}: Evaluate Q$_\delta$ statistic every 5 epochs and save checkpoints at optimal points
\end{enumerate}

\textbf{Generation Process:}
\begin{enumerate}
    \item Initialize with beginning-of-sequence token [BOS]
    \item For each position in the sequence:
    \begin{itemize}
        \item Feed current sequence through the model
        \item Apply column-aware token constraints to ensure valid domain values
        \item Sample next token from the constrained output distribution
    \end{itemize}
    \item Continue until end-of-sequence token [EOS] is generated
    \item Process generated token sequences through inverse data transformer
    \item Reconstruct numerical values from partitioned tokens and convert categorical tokens to original representations
\end{enumerate}

\textbf{Implementation Details:}
\begin{itemize}
    \item \textbf{Training Configuration}: Batch size 64 with gradient accumulation steps of 4 (effective batch size 256), learning rate $2 \times 10^{-4}$ with AdamW optimizer
    \item \textbf{Tokenization Strategy}: Single-character partitioning for numerical values provides fine-grained control during generation
    \item \textbf{Constrained Generation}: Column-aware token filtering prevents generation of invalid values (e.g., non-existent airport codes)
    \item \textbf{Early Stopping}: Bootstrap-based thresholds with patience of 2 consecutive evaluations above threshold
\end{itemize}

The architecture's strength lies in its ability to model tabular data as sequences, leveraging the proven capabilities of transformer architectures while incorporating domain-specific constraints and regularization techniques to ensure high-quality synthetic flight data generation.

\end{document}